\documentclass[12pt]{iopart}
\expandafter\let\csname equation*\endcsname\relax
\expandafter\let\csname endequation*\endcsname\relax
\usepackage{amssymb,amsmath}

\usepackage{hyperref}
\usepackage{multirow}
\usepackage{comment}
\usepackage{array}
\usepackage[acronym,toc,shortcuts]{glossaries}
\usepackage[usenames, dvipsnames]{color}
\usepackage{url}
\usepackage{csquotes}
\DeclareMathOperator*{\argmax}{arg\,max}

\usepackage{float}
\usepackage[acronym,toc,shortcuts]{glossaries}
\usepackage{graphicx}
\usepackage{caption}
\usepackage{subcaption}
\usepackage{soul}
\usepackage{algorithm}
\usepackage{algorithmic} 
\usepackage{enumitem}
\usepackage{booktabs,tabularx}
\usepackage{bbm}
\usepackage{amssymb}
\usepackage{amsmath}
\usepackage{upgreek}
\usepackage{arydshln}
\usepackage{array,tabularx} 
\newcolumntype{L}[1]{>{\raggedright\arraybackslash\hspace{0pt}}p{#1}}

\newcommand{\matr}[1]{\mathbf{#1}}
\begin{document}

\title[]{Source-Free Domain Adaptation for SSVEP-based Brain-Computer Interfaces}

\author{Osman Berke Guney$^1$, Deniz Kucukahmetler$^2$,
Huseyin Ozkan$^3$}

\address{$^1$ Department of Electrical and Computer Engineering, Boston University, Boston, MA, USA.}
\address{$^2$ Max Planck Institute for Human Cognitive and Brain Sciences, Leipzig, Germany.}
\address{$^3$ Faculty of Engineering and Natural Sciences, Sabanci University, Istanbul, Turkey.}
\ead{berke@bu.edu}
\vspace{10pt}

\begin{abstract}

\textit{Objective:} SSVEP-based BCI spellers assist individuals experiencing speech difficulties by enabling them to communicate at a fast rate. However, achieving a high information transfer rate (ITR) in most prominent methods requires an extensive calibration period before using the system, leading to discomfort for new users. We address this issue by proposing a novel method that adapts a powerful deep neural network (DNN) pre-trained on data from source domains (data from former users or participants of previous experiments), to the new user (target domain) using only unlabeled target data. \textit{Approach:} Our method adapts the pre-trained DNN to the new user by minimizing our proposed custom loss function composed of self-adaptation and local-regularity terms. The self-adaptation term uses the pseudo-label strategy, while the novel local-regularity term exploits the data structure and forces the DNN to assign similar labels to adjacent instances. \textit{ Main results:} Our method achieves excellent ITRs of \textit{201.15 bits/min} and \textit{145.02 bits/min} on the benchmark and BETA datasets, respectively, and outperforms the state-of-the-art alternatives. Our code is available at
\url{https://github.com/osmanberke/SFDA-SSVEP-BCI} \textit{Significance:} The proposed method prioritizes user comfort by removing the burden of calibration while maintaining an excellent character identification accuracy and ITR. Because of these attributes, our approach could significantly accelerate the adoption of BCI systems into everyday life.
\end{abstract}
\noindent{\it Keywords}: SSVEP, BCI, speller, deep learning, domain adaptation, domain generalization.

%

%
%
%
%

\section{Introduction}

\label{sec:IN}
Brain-computer interface (BCI) speller systems provide a means of communication for individuals who experience speech difficulties \cite{journal_bci}. Owing to its non-invasive nature and practical usability, electroencephalography (EEG) is a typical choice for measuring brain signals in these systems \cite{journal_eeg}. Steady-state visually evoked potential (SSVEP) is a brain response elicited by a visual object flickering at a certain frequency. Harmonics of the flickering frequency characterize the SSVEP response and can be observed in EEG signals \cite{ournetwork}. Because of this characteristic, SSVEP-based stimulation paradigms find substantial use in BCI speller systems \cite{ssvep_bci_review} and are known to yield impressive speller performance in terms of character identification (i.e., prediction) accuracy and information transfer rate (ITR) \cite{ssvep_common}. In these SSVEP-based BCI speller systems, the user chooses a character to be spelled from a collection of visuals flickering (each with a unique flickering frequency) on a computer screen and attends by gazing. At run time, the chosen character (the one intended to be spelled) can be identified by EEG signal classification. The goal is to maximize the ITR, enabling accurate character identification with relatively short stimulation \cite{ournetwork}. Since EEG signal statistics vary significantly from one person to another, existing character identification algorithms with high ITRs (such as \cite{ournetwork,convca}) require a quite tiring calibration process for each new user. This process must be conducted before actual system use begins and includes new EEG experiments, collecting labeled data, preprocessing, and algorithm training.  As the typical user is a disabled individual, it is of immense importance to remove this burden of calibration for user comfort and allow an immediate plug-and-play system start. 

To that end, transfer learning based methods \cite{our_ensemble,tt-CCA} aim to domain generalization across participants\footnote{The term user refers to the actual end user of the system and the term participants refer to individuals who participated in previously conducted EEG experiments or individuals who are prior users. Moreover, similar to the machine learning literature, we refer to the participants as source domains and to the user as target domain.} for knowledge extension to the new users. One example can be found in \cite{our_ensemble}, which trains/develops algorithms using previously collected data of participants (source domains) and directly transfers to the new user (target domain) as is, without requiring a calibration process. Although these methods provide user comfort, their character identification performances (i.e. prediction accuracy) are still far behind the best-performing methods (those requiring labeled data from the target domain). On the other hand, one can consider source-free domain adaptation (SFDA) \cite{SFDA_EEG_privacy,SFDA_EEG2} to achieve both user comfort and satisfactory identification performance. SFDA based approaches adapt the transferred method (originally trained with labeled data from source domains) to the target domain by using only unlabeled data from the target domain. Therefore, their (ones using unlabeled target data) performances are generally better than the domain generalization based ones (using no target data) of direct transfer. Since SFDA approaches do not store and, in fact, discard the data from source domains after training, they are called ``source-free". Moreover, SFDA does not compromise the user comfort as it does not require a separate calibration or labeling process for the new user. Notice that the SFDA framework is a suitable and natural choice for SSVEP-based BCI speller systems. As a new user interacts with the system, their unlabeled data continually accumulates and can be used for adapting the transferred method to gradually improve the initial transfer performance. Regarding the labeled data from source domains, it is typically available in large amounts and collected from offline past experiments with various healthy or disabled participants. One can even use the publicly available literature datasets for this purpose, such as benchmark \cite{benchmark_Dataset} and BETA \cite{beta}.

\begin{figure*}[t!]
\centering
\includegraphics[width=0.8\textwidth]{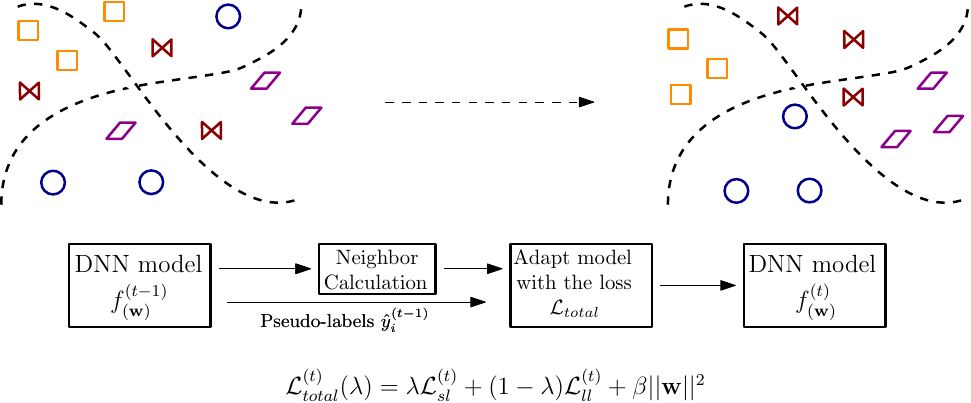}
\centering
\caption{{The outline of our source-free domain adaptation method and a representative illustration of changes in the DNN's decision boundaries across two consecutive iterations of adaptation are shown.  The left panel depicts the state before adaptation, and the right panel shows the state after adaptation. Dashed lines represent decision boundaries, while different shapes denote different classes.}
The adaptation starts with a DNN pre-trained using data from source domains, and it is carried out by minimizing the loss function $\mathcal{L}_{{total}}$ that we also introduce, which consists of two main terms: self-adaptation $\mathcal{L}_{sl}$ and local-regularity $\mathcal{L}_{{ll}}$. The self-adaptation component utilizes the pseudo-labeling approach, whereas the local-regularity component leverages the inherent structure of the data, forcing the adapted DNN to give similar labels to adjacent instances. Our method is compatible with any DNN architecture and particularly achieves ITR results of \textit{201.15 bits/min} and \textit{145.02 bits/min} on the benchmark \cite{benchmark_Dataset} and BETA \cite{beta} datasets, respectively, when utilizing the DNN architecture from \cite{ournetwork}. These results show that our proposed method provides significant ITR improvements over the state-of-the-art alternatives.}
\label{fig:adaptation_poster}
\end{figure*} 

In this paper, in order to prioritize user comfort while also achieving decent performance, we propose a novel SFDA algorithm for character identification in SSVEP-based BCI speller systems. Our algorithm adapts a previously trained (using source domains' labeled data) deep neural network (DNN) to the target domain (new user) based solely on unlabeled data from the target domain. Additionally, in our method, there is no need to store the data of source domains and so we discard it after training the initial DNN. This helps us save memory and prevent any privacy concerns that can perhaps arise due to data being transferred from the past participants \cite{SFDA_EEG_privacy}. At the heart of our algorithm is a novel loss function that we introduce for the adaptation process, which consists of two terms: one is a self-adaptation loss $\mathcal{L}_{sl}$, and the other is a local-regularity loss $\mathcal{L}_{{ll}}$ (along with standard L2 regularization). We compare our method with the strongest state-of-the-art alternatives \cite{oacca,comb-tCCA,tt-CCA} on two widely used publicly available datasets, benchmark \cite{benchmark_Dataset} and BETA \cite{beta}, and show that our method statistically significantly outperforms the others in terms of both accuracy and ITR. 

\section{Related Work} \label{sec:related_work}
Character identification methods in the SSVEP-based BCI speller literature can be divided into two groups based on their calibration requirements for new users: user-dependent and user-independent.  The user-dependent methods, such as {\cite{ournetwork,convca,trca,r3_p3}}, require a long and tiring calibration/preparation process (before actual system use) to collect labeled data from each new user, whereas the user-independent methods, such as \cite{CCA_method,fbcca}, do not. User-dependent methods generally exhibit superior performance, but that is at the cost of calibration discomfort \cite{to_train} which {is considered to hinder } the widespread use of SSVEP-based BCI speller systems. As we believe that the user comfort must not be compromised, we opt in this study for a method falling in the second group of user-independent methods. Our related work discussion here focuses on this second group, and we refer interested readers to \cite{to_train,ournetwork} for detailed information about the first group of user-dependent methods.

The group of user-independent methods can be further divided into three categories, which are (i) completely training-free methods, (ii) domain generalization and (iii) domain adaptation based methods. Here, we use machine learning terms to stimulate novel perspectives for SSVEP character identification by adopting from the literature of domain generalization and adaptation. 

{\bf Completely training-free methods} rely on mathematical models of the SSVEP signal characteristics.  A frequently employed method in the literature \cite{CCA_method} utilizes canonical correlation analysis (CCA) \cite{cca_statistic} to calculate the correlation coefficients between multidimensional reference signals and the received multi-channel EEG signal. Those reference signals are formed for each character, consisting of sinusoids at the harmonics of corresponding flickering frequency. CCA identifies the combination of harmonics and channels that maximize the correlation coefficient between each reference signal and the EEG signal \cite{CCA_method}. Then the character whose reference signal has the maximum correlation coefficient is declared as the spelled character. As in \cite{ournetwork}, we refer to this method as Standard-CCA throughout the paper. A filter-bank extension of Standard-CCA (FBCCA) is proposed in \cite{fbcca}. In FBCCA, multiple band-pass filters with different low cut-off frequencies first preprocess the input EEG, and Standard-CCA is then applied to each preprocessed signal. The intended character is predicted based on the weighted combination of calculated correlation coefficients. FBCCA has been demonstrated to achieve higher accuracy and ITR than Standard-CCA \cite{fbcca}. The advantage of these methods is that they are usable in a plug-and-play manner since they do not require any form of training. However, because of the same reason, they underperform, especially in settings with short stimulation durations.

{\bf Domain generalization based methods} leverage data from source domains to train/formulate models/templates that are directly applied in the target domain. For example, transfer template CCA (tt-CCA) \cite{tt-CCA} averages the data from source domains to form template signals for each character. Along with the correlation coefficient calculated as in the Standard-CCA method, tt-CCA calculates two additional correlation coefficients using the formed template signals and predicts the spelled character based on the combination of these three. The tt-CCA method's accuracy and ITR performances are superior to the Standard-CCA method \cite{tt-CCA}. Another domain generalization based method, combined-CCA (C3A) \cite{comb-tCCA}, predicts the spelled character in a manner similar to tt-CCA, but with a different calculation of the correlation coefficients. In \cite{our_ensemble}, the DNN architecture of \cite{ournetwork} is first trained for each source domain. The resulting ensemble of DNNs is then transferred to the target domain, where the most representative DNNs from the top $k$ source domains are used for character prediction. {At the time of publication, this ensemble approach achieved the highest performance among domain generalization-based methods \cite{our_ensemble}.  More recently, CCA-Net \cite{cca_net}, a method that combines tt-CCA \cite{tt-CCA} with the approach from \cite{osssvep}, has been proposed and demonstrates a slight performance improvement over the ensemble of DNNs.} This category of methods is certainly practical as they do not require any labeled data from the target domain. However, since they do not make any sort of adaptation to the target domain, their performances are not as satisfying as the group of user-dependent or the category of domain adaptation (but generally better than the category of completely training-free). Also note that the domain adaptation based methods are usually built on top of the methods from this category, as explained next. Thus, developing domain generalization based methods are equally important as developing domain adaptation based ones.

{\bf Domain adaptation based methods} adapt to the target domain in an unsupervised fashion. One such method is online tt-CCA (ott-CCA), which extends tt-CCA by updating the transferred template signals using unlabeled target domain data. It has been shown that ott-CCA improves the performance of tt-CCA  \cite{tt-CCA}. Similarly, adaptive-C3A updates the transferred template signals in the C3A method, which also yields a considerable performance enhancement \cite{comb-tCCA}. A slightly different method, called online adaptive CCA (OACCA) \cite{oacca}, adapts FBCCA via new user-specific channel and harmonic combinations that are calculated based on techniques from \cite{PSF} and \cite{msCCA}. We observe that to date, OACCA is the best-performing method in this category. Our preliminary simple pseudo-labeling adaptation, presented in a short conference proceeding (in Turkish) \cite{siu_adaptation}, also falls in this category. In that proceeding, we adapt the DNN architecture of \cite{ournetwork} to the target domain by using only the self-adaptation term $\mathcal{L}_{sl}$. The present work, however, introduces several enhancements and novel features (such as the local-regularity term $\mathcal{L}_{{ll}}$) as detailed in Section \ref{sec:novel}. We compare our proposed method against the strongest alternatives OACCA \cite{oacca}, adaptive-C3A \cite{comb-tCCA}, and ott-CCA \cite{tt-CCA} methods. The experimental results show that our method delivers much higher identification accuracy and ITR figures, outperforming them altogether.

We also briefly highlight here a few related source-free domain adaptation methods from the machine/deep learning literature. A method referred to as SHOT, proposed in \cite{shot_sfda}, adapts a network to the target domain. It freezes the network's last layer (the classifier part) and learns target-domain-specific features by adapting the remaining parts of the network (the feature extractor part) using pseudo-labeling. In \cite{affinity_sfda}, the authors propose a method called neighborhood reciprocity clustering (NRC) that exploits an intrinsic data structure and attempts to cluster similar instances. In another study \cite{lln_sfda}, the authors approach the SFDA from the label noise perspective, where they use a regularizer to mitigate the label noise memorization problem. These studies inspired us in a very general sense. We used similar -but not the same- ideas in our adaptation method. For instance, our local-regularity loss term also exploits the structure of data, but our term is entirely different from the NRC method \cite{lln_sfda} and is designed to specifically utilize the characteristics of SSVEP signal. The SFDA literature is rapidly expanding; and we believe that in future studies, new ideas could be incorporated with our novel adaptation features here to generate more versatile machine/deep learning techniques.

\subsection{Novel Contributions and Highlights}\label{sec:novel}
{In this paper, we propose a novel source-free unsupervised domain adaptation method that is compatible with any architecture. Our approach adapts a pre-trained DNN to the target domain, prioritizing practicality and user comfort in SSVEP-based BCI speller systems.} The adaptation is carried out by minimizing our proposed custom loss function, composed of self-adaptation $\mathcal{L}_{sl}$ and local-regularity $\mathcal{L}_{{ll}}$ terms, based only on unlabeled data from the target domain. We evaluate the performance of our method on two publicly available benchmark \cite{benchmark_Dataset} and BETA \cite{beta} datasets. Our method demonstrates superior performance compared to the current state-of-the-art approaches, achieving an ITR of $201.15$ bits/min on the benchmark dataset \cite{benchmark_Dataset} and $145.02$ bits/min on BETA, using the DNN architecture from \cite{ournetwork}. Our code is available for reproducibility at

\noindent
\url{https://github.com/osmanberke/SFDA-SSVEP-BCI}

\begin{itemize}
    \item In our method, new users can immediately start using the system in a plug-and-play manner since there is no need for a calibration process to conduct tiring and long additional EEG experiments, pre-processing and algorithm training. {In this sense, we prioritize user comfort in particular}. Also, the system performance continuously improves by adapting to the unlabeled data accumulated by the new user as s/he interacts with the system. Hence, we believe that our method has the potential to promote the widespread use of SSVEP-based BCI applications, such as gaming \cite{gaming1}, in daily lives.
    
    \item To the best of our knowledge, this study presents (along with our preliminary short proceeding \cite{siu_adaptation}) the first DNN based domain adaptation method in the SSVEP character identification literature. It also achieves the highest ITR in its own group of user-independent methods.
    
    \item A notable novelty in our work is the local-regularity term $\mathcal{L}_{{ll}}$ in the loss function that we introduce specifically for the SSVEP adaptation process. This term exploits the underlying data structure and forces the adapted DNN model $f_\matr{w}$ to give similar labels to neighbouring instances. It is worth noting that our regularity term here is technically completely different from that of the NRC method \cite{affinity_sfda}.
    
    \item Furthermore, regarding local-regularity, we present a distinctive and innovative strategy for determining the set of neighbors dynamically (allowing a variable number of neighbors) for every SSVEP instance since (i) the user may spell some characters more often than others and (ii) the noise level/EEG signal statistics can change from one domain (person) to another or even temporally in the same domain (for the same person).
    
    \item Our method does also dynamically optimize the weights of the two loss terms, controlled by a $\lambda$ parameter, for each target domain (new user) separately, since the degree of statistical similarity with the source domains (participants) can be quite different for each target domain. 
\end{itemize}
\section{Problem Description}
In SSVEP-based EEG BCI speller systems, the spelled character is predicted from $M$ possibilities $y \in \{1,2,\cdots,M\}$. The prediction is based on the multi-channel EEG signal $\matr{x}\in\mathbb{R}^{C\times N_T}$ received during the stimulation\footnote{We use the plain math type notation for scalar variables but the bold text type for vectors and matrices. For example, the multi-channel EEG signal is shown as $\matr{x}$ instead of $x$.}, where $C$ is the number of channels (EEG electrodes), $N_T=T\times F_s$ is the signal length, $T$ (sec) is the stimulation duration and $F_s$ is the sampling frequency (Hz). The goal is to maximize ITR \cite{itr_cite}, 

\begingroup
\begin{equation*}
    \text{ITR}(P,T) = \left( \log_{2} M + P\log_{2} P + (1-P)\log_{2} \left( \frac{1-P}{M-1} \right) \right) \frac{60}{T}
\end{equation*}
\endgroup

\noindent \noindent 
which can be achieved by optimizing two incompatible objectives: maximizing the prediction accuracy $P$ and minimizing the stimulation duration $T$ \cite{ournetwork}. Note that $P$ and $T$ are strongly coupled: When data processing is optimal, the only way to improve accuracy (higher $P$) is to observe more data (longer $T$), which does not necessarily translate into a higher ITR.  

A viable strategy here is to maximize the prediction accuracy for a fixed $T$, yielding $P^*(T)$, and choose $T^*=\argmax_{T} \text{ITR}(P^*(T),T)$ as the optimal stimulation duration \cite{ournetwork}. For a fixed stimulation duration $T$, the task of maximizing the prediction accuracy $P$ can be studied as a multi-class classification problem, i.e., character identification.

\section{Method}  \label{sec:methodology}

We constrain the above-described multi-class classification problem with user comfort as explained in Section \ref{sec:IN}. Particularly, our aim is to remove the burden of the calibration process from the user and allow an immediate plug-and-play system start. Therefore, unsupervised and source-free domain adaptation is a viable approach,  where unlabeled data from the target domain is exploited.  

Before we explain the details, let us assume that a set of labeled EEG data instances from a collection of participants (source domains) is available: $\{\{(\matr{x}_i^l,y_i^l)\}_{i=1}^{D_l}\}_{l=1}^{N_{tr}}$, where $N_{tr}$ is the number of source domains, and $D_l$ is the total number of EEG instances {from the $l$-th participant}. We also use $\matr{z}^l_i \in \{0,1\}^{M\times 1}$ as {the one-hot encoding} of label $y^l_i$ such that $\matr{z}^l_{i,y^l_i}=1$, where $\matr{z}^l_{i,y^l_i}$ is the $y^l_i$-th element of $\matr{z}^l_i$, and $0$ otherwise. Additionally, a set of unlabeled EEG data instances from the new user (target domain) is denoted by $\{\matr{x}_i\}_{i=1}^{N}$. 

{We start by transferring a deep neural network $f_\matr{w}$ that is pre-trained with data $\{\{(\matr{x}_i^l,y_i^l)\}_{i=1}^{D_l}\}_{l=1}^{N_{tr}}$ from the source domains to the target domain, yielding a high enough initial accuracy.} Then, we adapt the network $f_\matr{w}$ ($\matr{w}$ keeps the network parameters) to the target domain with unlabeled data generated during the process as the user (target domain) interacts with the system. Thus, the initial transfer accuracy improves through the introduced unsupervised adaptation. The network $f_\matr{w}$ gives a soft response $f_\matr{w}(\matr{x}_i)=\matr{s}_i\in[0,1]^{M\times1}$ to the input $\matr{x}_i$ with the prediction $\hat{y}_i$ being the index of the maximum in $\matr{s}_i$, i.e, $\hat{y}_i = \argmax_j \matr{s}_{i,j}$.

Our method is mainly based on a novel loss function that we introduce for the unsupervised adaptation of the transferred network $f_\matr{w}$. This introduced loss, which is minimized using unlabeled data from the target domain with expectation-maximization (EM) type of iterations \cite{EM_algorithm}, consists of two terms: self-adaptation $\mathcal{L}_{sl}$ and local-regularity $\mathcal{L}_{{ll}}$ terms.

{\bf Self-Adaptation Loss $\mathcal{L}_{sl}$: } In the presence of true target data labels $y_i$, the model $f_\matr{w}$ would be adapted to the target domain by minimizing the standard cross entropy loss:
\begin{align*}
&-\frac{1}{N}\sum_{i=1}^{N}\sum_{j=1}^{M} \matr{z}_{i,j}\log(\matr{s}_{i,j})=-\frac{1}{N}\sum_{i=1}^{N} \log(\matr{s}_{i,y_i}),
\end{align*}
where the equality follows because the $y_i$-th index of the one-hot encoding vector $\matr{z}_i$ equals to $1$, and the other indexes are $0$ (with $0\log0=0$).

However, since we do not have the true labels in the SFDA setting, predictions $\hat{y}_i$ of the model can be used as pseudo-labels for adaptation (hence the model $f_\matr{w}$ is updated) which -in turn- yields new predictions and a new adaptation. This defines a single iteration which we repeat until convergence. Let $\matr{w}^{(t)}$ be the adapted parameters and let $\matr{s}_i^{(t)}$ and $y_i^{(t)}$ be the network soft responses and predictions, respectively, for the input $\matr{x}_i$ at the end of iteration $t$, i.e, $\matr{s}_i^{(t)} = f_{\matr{w}^{(t)}}(\matr{x}_i)$ and $\hat{y}_i^{(t)} = \argmax_j \matr{s}_{i,j}^{(t)}$. Within this scheme, such EM type of iterations define our first term of self-adaptation loss as

\begin{equation}
    \mathcal{L}_{sl}^{(t)} = -\frac{1}{N}\sum_{i=1}^{N} \log(\matr{s}_{i,\hat{y}_i^{(t)}}^{(t)}). \label{eq:self_adaptation_term}
\end{equation}

The efficacy of minimizing loss functions using pseudo-labels has been previously shown in many unsupervised \cite{cvpr_uda_pseudo} and semi-supervised \cite{pseudo_ssl,entropy_regularization,text_ssl_pseudo} settings. In effect, the network learns classes that are well separated in low-density regions by minimizing the class overlaps \cite{pseudo_ssl}. It is also equivalent to entropy regularization \cite{entropy_regularization}, which aims to benefit from unlabeled data in the maximum a posterior sense. Another perspective can be taken from the concept of dropout regularization, proposed in \cite{dropout_paper} to reduce overfitting. When the dropout layer is used in any model, in the phase of training, a different part (thinned version) of the network is used every time it forms a response. Whereas in the phase of testing, responses are always formed based on the full network (no node dropping). Responses of full network are generally better (having less errors) than those of a partial network, cf. Fig. 11 in \cite{dropout_paper}. {Namely, the full network $f_{\matr{w}^{(t)}}$ is used when generating the pseudo-labels and partial networks derived from $f_{\matr{w}^{(t)}}$ with node dropping are used when minimizing the self adaptation loss in \eqref{eq:self_adaptation_term}. Hence, the quality difference between the responses of the model obtained during training and testing can explain performance gains in using pseudo-labels. Additionally, it is worth noting that prominent DNN architectures in the SSVEP classification literature, such as \cite{ournetwork, ssvepformer}, typically incorporate dropout layers. }

Although minimizing self-adaptation loss is effective, it does not take into account the structure of the available unlabeled data from the target domain. We next explain our proposed local-regularity loss $\mathcal{L}_{{ll}}$ that exploits the data structure.

{\bf Local-Regularity Loss $\mathcal{L}_{{ll}}$:} The idea behind local-regularity is that close (or similar) enough data instances should attain similar labels (Fig. \ref{fig:adaptation_poster}), which has been successfully exploited in many past studies of algorithm design. A couple of examples can be found in source-free domain adaptation \cite{shot_sfda,affinity_sfda}, metric learning \cite{kulis2012metric,massimino2018like}, and kernel learning \cite{gonen2011multiple}. Accordingly, we use the correlation coefficient\footnote{Since the SSVEP instances we consider here are multidimensional ($\matr{x}_i$'s are in the matrix form of dimension $C \times N_T$), we calculate the correlation coefficient $\rho(\matr{x}_i,\matr{x}_j)$ after applying a common channel combination $\matr{w}_{c}$ to both instances. The details of selecting the channel combination and calculations are explained in Appendix.} as a measure to define closeness (similarity) and obtain our local-regularity loss $\mathcal{L}_{{ll}}$ as
\begin{equation}
   \mathcal{L}_{{ll}}^{(t)} =-\frac{1}{N}\sum_{i=1}^{N}\frac{1}{k_i}\sum_{j=1}^{k_i} \log(\matr{s}_{i,\hat{y}_{I_i(j)}^{(t)}}^{(t)}), \text{ where }
   \label{eq:local_regularity_term}
\end{equation}

\begin{itemize}
\item $I_i$ is the set of indices that are sorted in descending order based on the correlation coefficient values between the target domain instances $\{\matr{x}_j\}_{j=1,j\neq i}^{N}$ and the target domain instance $\matr{x}_i$. Namely, $\matr{x}_{I_i(1)}$ is the most correlated (closest) to $\matr{x}_i$.
\item $k_i$ is the number of neighbors considered for the instance $\matr{x}_i$. Note that the neighborhood size is not fixed, and is allowed to vary across instances. This is because the user might well spell certain characters more often than others, or the neighborhood size (set of similar instances) can shrink or expand depending on the noise level, which can vary from domain to domain or even from instance to instance in the same domain.
\item Superscript $^{(t)}$ denotes the iterations as described earlier.
\end{itemize}

The minimization of this loss enforces the transferred network $f_{\matr{w}^{(t)}}$ to assign similar predictions to instances that are in close proximity. Note that the correlation coefficient value is inversely proportional to the squared Euclidean distance after centering and unit-norm normalization. In fact, most of the methods explained in Section \ref{sec:related_work} such as \cite{our_ensemble,oacca,fbcca} use the correlation coefficient as a suitable metric for measuring similarities between SSVEP signals.

As for determining the variable neighborhood size $k_i$ properly, it is reasonable to assume that the correlation coefficients between the instance $\matr{x}_i$ and its neighboring instances that should share similar labels with $\matr{x}_i$ are much higher compared to the other instances. As a result, one could expect a significant drop from the $k_i$'{th} highest correlation coefficient to the next $k_i+1$'{th} highest:
\begingroup
\begin{equation}
    \frac{\rho(\matr{x}_i,\matr{x}_{I_i(k_i)})-\rho(\matr{x}_i,\matr{x}_{I_i(k_i+1)})}{|\rho(\matr{x}_i,\matr{x}_{I_i(k_i)})|} \geq  \delta, \label{eq:threshold_neighbour}
\end{equation}
\endgroup
where the left hand side of the inequality describes the percentage drop and the right hand side $\delta$ is a predetermined threshold testing the significance of the drop. The neighborhood size $k_i$ is taken from the point of significant drop.

{\bf Total Loss: } The combination of the self-adaptation \eqref{eq:self_adaptation_term} and local-regularity \eqref{eq:local_regularity_term} terms, along with an L2 regularization {weighted by $\beta$}, yields the total loss:
\begin{equation}
    \mathcal{L}_{{total}}^{(t)}(\lambda)=\lambda\mathcal{L}_{sl}^{(t)}+(1-\lambda)\mathcal{L}_{{ll}}^{(t)} + \beta ||\mathbf w||^{2},
    \label{eq:total_loss}
\end{equation}
where $\lambda\in[0,1]$ controls the contribution of individual loss terms.

Furthermore, a dynamic parameter selection procedure is proposed for $\lambda$ as follows. Firstly, we adapt the DNN $f_{\matr{w}^{(t-1)}}$ to the target domain by minimizing $\mathcal{L}_{{total}}^{(t-1)}(\lambda)$ separately for every value from a set of candidates $\lambda \in \Lambda$, yielding $\matr{w}^{(t)}(\lambda)$ for the adapted parameters and $\hat{y}^{(t)}_i(\lambda)$ for the predictions. Then, to decide which of the adapted network parameters $\{\matr{w}^{(t)}(\lambda): \lambda \in \Lambda$\} is to be used for a final prediction (either at the end of iterations or intermediately if needed), we check how well each adapted parameters $\matr{w}^{(t)}(\lambda)$ cluster the unlabeled target domain data. The silhouette clustering metric \cite{silhouette} is used for that purpose, which generates a score $m_i^{(t)}({\lambda})$ to assess how well an instance $\matr{x}_i$ is clustered based on the difference between the tightness $a_i^{(t)}({\lambda})$ of $\matr{x}_i$ to its own cluster and separation $b_i^{(t)}({\lambda})$ of $\matr{x}_i$ from other clusters: 

\begin{align}
    a_i^{(t)}({\lambda}) &={\frac{1}{q^{(t)}_{\hat{y}_{i}^{(t)}}({\lambda})-1}} \sum_{\substack{j=1 \\ j\neq i}}^{N} d(\matr{x}_j,\matr{x}_i)\mathbbm{1}(\hat{y}_{j}^{(t)}({\lambda})=\hat{y}_{i}^{(t)}({\lambda})), \nonumber \\
    b_i^{(t)}({\lambda}) &=\min_{\substack{k \in \{1,\cdots,M\} \\ \backslash \{\hat{y}_{i}^{(t)}({\lambda})\}}} \frac{1}{q_k^{(t)}(\lambda)} \sum_{\substack{j=1 \\ j\neq i}}^{N} d(\matr{x}_j,\matr{x}_i)\mathbbm{1}(\hat{y}^{(t)}_{j}({\lambda})=k), \nonumber \\
    m_i^{(t)}({\lambda}) &= \frac{b_i^{(t)}({\lambda})-a_i^{(t)}({\lambda})}{\max(a_i^{(t)}({\lambda}),b_i^{(t)}({\lambda}))}, \text{where}
    \label{eq:silhouette}
\end{align}

\begin{itemize}
\item $\mathbbm{1}(.)$ is the indicator function that returns $1$ if its condition is satisfied (otherwise $0$), $q^{(t)}_k({\lambda})$ is the total number of instances labeled as $k$ by the DNN with parameters $\matr{w}^{(t)}({\lambda})$, i.e., $q^{(t)}_k({\lambda}) = \sum_{j=1}^{N} \mathbbm{1}(\hat{y}^{(t)}_{j}({\lambda})=k)$,

\item $d(\matr{x}_i,\matr{x}_j)$ is a distance metric derived from the correlation coefficient as $d(\matr{x}_j,\matr{x}_i)=1-\rho(\matr{x}_j,\matr{x}_i)$, which is equivalent to cosine distance if $\matr{x}_i$'s are zero mean signals,

\item $a^{(t)}_i({\lambda})$ is the average distance from $\matr{x}_i$ to the other instances of the same predicted label, i.e., $a^{(t)}_i({\lambda})$ equals $1$ if  $q^{(t)}_{\hat{y}_{i}^{(t)}}({\lambda})=1$,

\item $b^{(t)}_i({\lambda})$ is the minimum average distance from $\matr{x}_i$ to the instances of another cluster (of a different predicted label) with the minimization being over all other clusters,

\item the superscript $^{(t)}$ denotes the iterations as described earlier.
\end{itemize}
The overall silhouette clustering score for the predictions of the model with parameters $\matr{w}^{(t)}(\lambda)$ is the average of the silhouette scores of all the instances, i.e., $m^{(t)}(\lambda) = \sum_{i=1}^{N} m_i^{(t)}(\lambda) /N$. The network parameters $\matr{w}^{(t)}(\lambda)$ with the highest silhouette clustering score $m^{(t)}(\lambda)$ is selected.

The adaptation is continued separately for each $\lambda$ until convergence that is controlled with the overall silhouette clustering score $m^{(t)}(\lambda)$. It is terminated, if the minimization of $\mathcal{L}_{{total}}^{(t)}(\lambda)$ does not yield a sufficient improvement in the clustering score compared to the previous iteration $m^{(t-1)}(\lambda)$ over $B$ consecutive trials. At each trial of an iteration $t$, the adaptation is started with the DNN's weights $\matr{w}^{(t-1)}$ of the previous iteration $t-1$. The reason for using these multiple trials (applied when the DNN contains dropout layers) is that the weights converge to a different state at each trial because of the randomness caused by the dropout layers. Also note that $J$ epochs are employed for stable training at each iteration.

Lastly, prediction confidences have been shown to be useful in the unsupervised learning literature \cite{cvpr_uda_pseudo}, \cite{confidence2}. We also take into account the confidence aspect in our method, because each instance is clustered with varying degrees of confidence; while some instances are clustered well, others are not. Therefore, we use the pseudo-labels of only those instances that are clustered well with positive silhouette values $m_i^\lambda$. The reason behind this approach is that an instance with a negative silhouette score implies that there exists a cluster closer to this instance than the cluster it is attributed to. Hence, instances with negative silhouette scores are more likely to be misclassified. To avoid updating the DNN using the pseudo-labels that are likely to be incorrect, we update with only the pseudo-labels of the instances having a positive silhouette score. We set $\delta=0.05$, $\beta=0.001$, $B=3$, $J=50$, and $\Lambda=\{0, 0.2, 0.4, 0.6, 0.8, 1\}$. All implementation details, including {the glossary of parameters (Glossary \ref{table:parameters})} and the algorithmic outline of our adaptation method (Algorithm \ref{alg:adaptation}), are provided in the Appendix.

\begin{figure}[t!]
    \centering
    \begin{subfigure}[b]{0.7\textwidth}
        \centering
        \includegraphics[width=\textwidth]{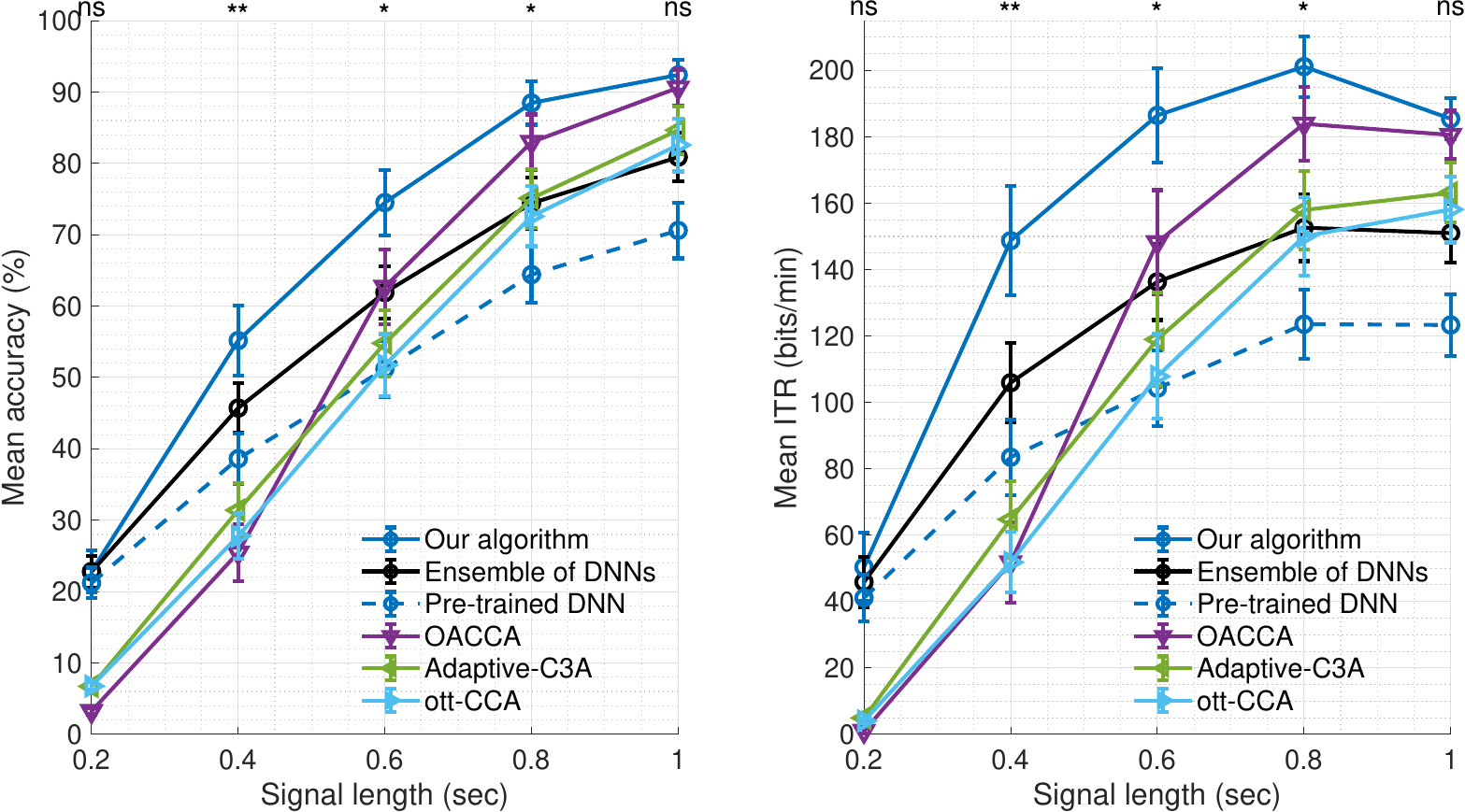}
        \caption{}
        \label{fig:bench_res}
    \end{subfigure}
    \hspace{0.05\textwidth}
    \begin{subfigure}[b]{0.7\textwidth}
        \centering
        \includegraphics[width=\textwidth]{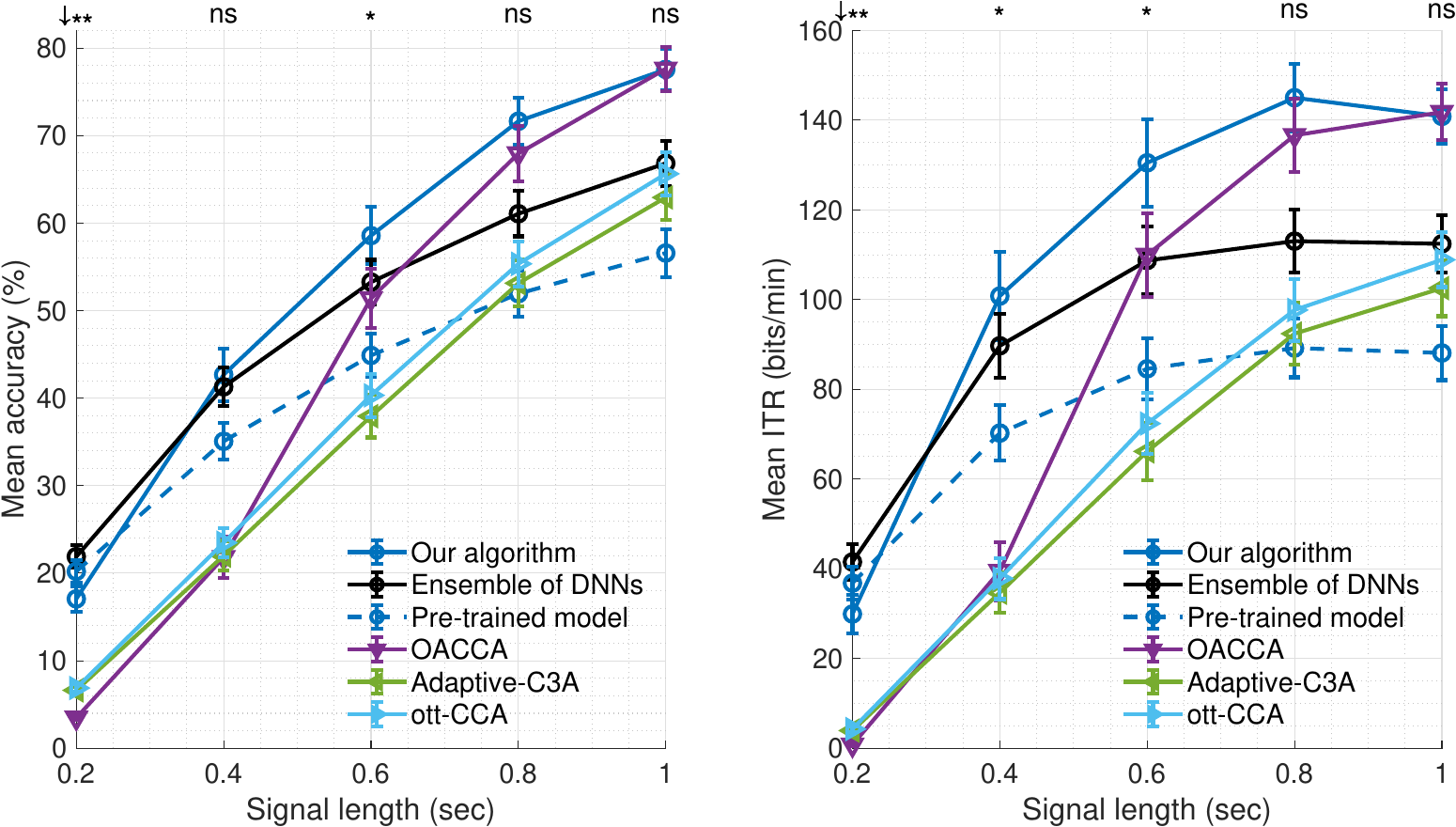}
        \caption{}
        \label{fig:beta_res}
    \end{subfigure}
    \caption{The mean classification accuracy on the left and the mean information transfer rate (ITR) on the right are presented across all users in the datasets, together with the standard errors indicated by the bars. The asterisks indicate the results of statistical significance analyses. Paired $t$-tests are applied, and significance levels are reported based on the least significant difference criterion (*$p < \frac{0.05}{4}$, **$p < \frac{0.05}{20}$). The notation `ns' denotes results that are not statistically significant. The marker $^{\downarrow}$ denotes effects favoring the comparator. (a) The results for the benchmark dataset \cite{benchmark_Dataset} with 35 users. (b) The results for the BETA dataset \cite{beta} with 70 users.}
    \label{fig:subfigures}
\end{figure}

\section{Performance Evaluations} \label{sec:evaluations}
We compare our source-free domain adaptation method with three state-of-the-art domain adaptation alternatives from the literature, OACCA \cite{oacca}, adaptive-C3A \cite{comb-tCCA} and ott-CCA \cite{tt-CCA}. {We also compare our method with one of the state-of-the-art domain generalization methods, the ensemble of DNNs \cite{our_ensemble}.} The presented comparisons are based on the large scale (EEG experiments with $105$ participants in total) benchmark \cite{benchmark_Dataset} and the BETA \cite{beta} datasets which are publicly available and widely used in the literature. The performance evaluation is conducted in a leave-one-participant-out fashion. One participant is reserved and considered as the user (target domain), while the remaining participants' data (source domains) are exploited to obtain the pre-trained DNN $f_\matr{w}$. We next adapt the DNN to the target domain in an unsupervised manner, using all the unlabeled data of the target domain, and calculate the accuracy and ITR performance with true labels at the end of the adaptation. This procedure is repeated for each participant, so we have $35$ and $70$ repetitions for the benchmark and BETA datasets, respectively. We evaluate the performance of our adaptation method using three architectures: the main DNN architecture \cite{ournetwork}, SSVEPformer \cite{ssvepformer}, and an alternative DNN architecture \cite{another_network}. Unless otherwise specified in the experiment results, we use the main DNN architecture proposed in \cite{ournetwork}, as it achieves the highest accuracy and ITR performance in the user-dependent setting on the benchmark \cite{benchmark_Dataset} and BETA \cite{beta} datasets. Detailed descriptions of these three architectures are provided in the Appendix.

The adaptation is continued until the silhouette clustering metric converges in our method. On the contrary, there is no convergence control in the compared methods since their solutions do not involve losses or convergences. To ensure a fair comparison, nonetheless, we apply the same procedure to the compared methods as well. Namely, the adaptation is continued until all the predictions remain the same for two successive iterations in the compared methods. The same input is given to all compared methods to receive their initial predictions, which are then used as pseudo-labels for updating their template/channel and harmonic combinations as explained in their respective papers. This resumes iteratively until the pseudo-labels are no longer updated in the new iteration, leading to a unique number of iterations for each method. Similarly, the performance is evaluated with true labels of the participant reserved as the target domain.

\begin{figure*}[t!]
    \centering
    \begin{subfigure}[b]{0.47\textwidth}
        \centering
        \includegraphics[width=\textwidth]{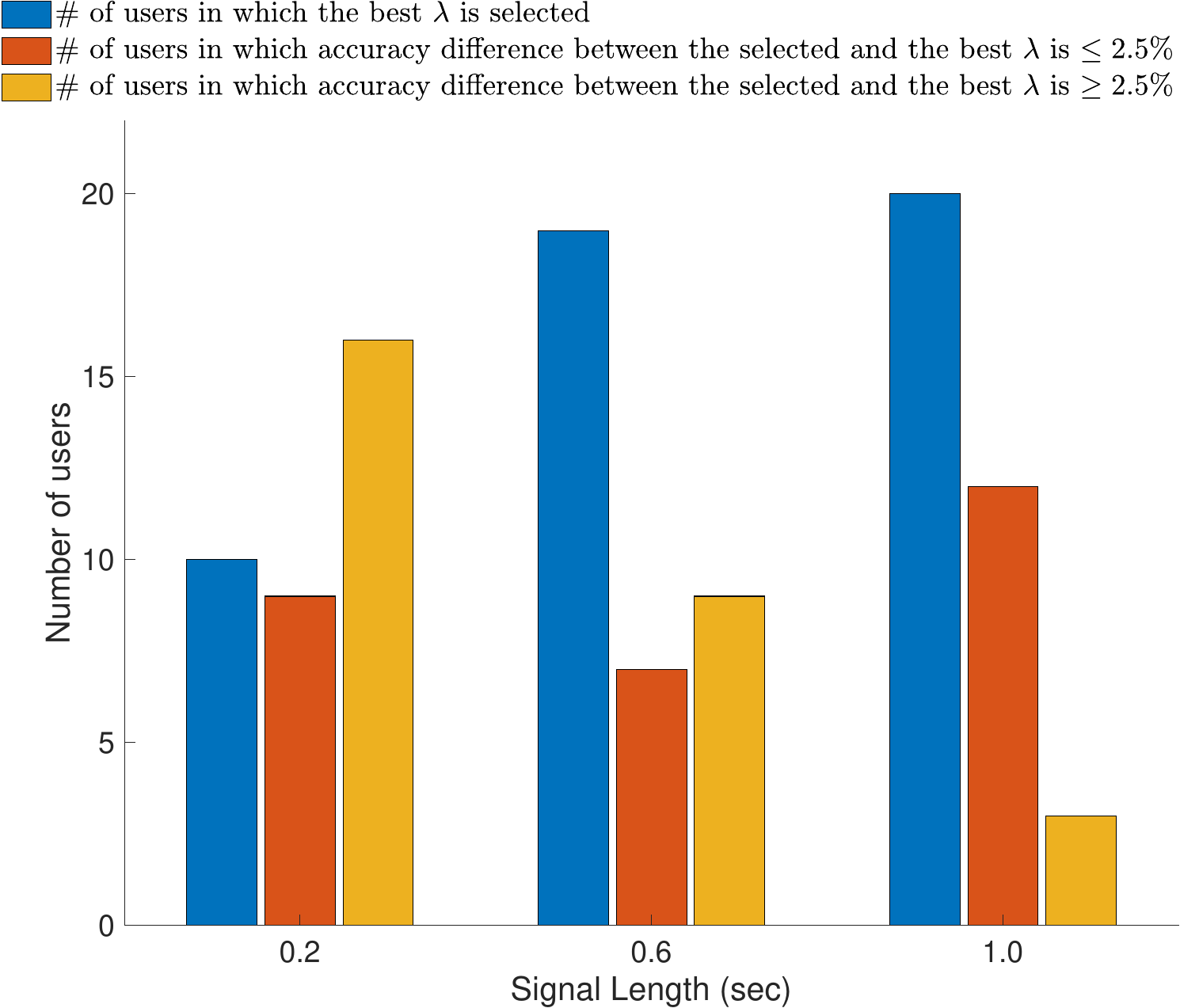}
        \caption{}
        \label{fig:best_lambda_bench}
    \end{subfigure}
    \begin{subfigure}[b]{0.47\textwidth}
        \centering
        \includegraphics[width=\textwidth]{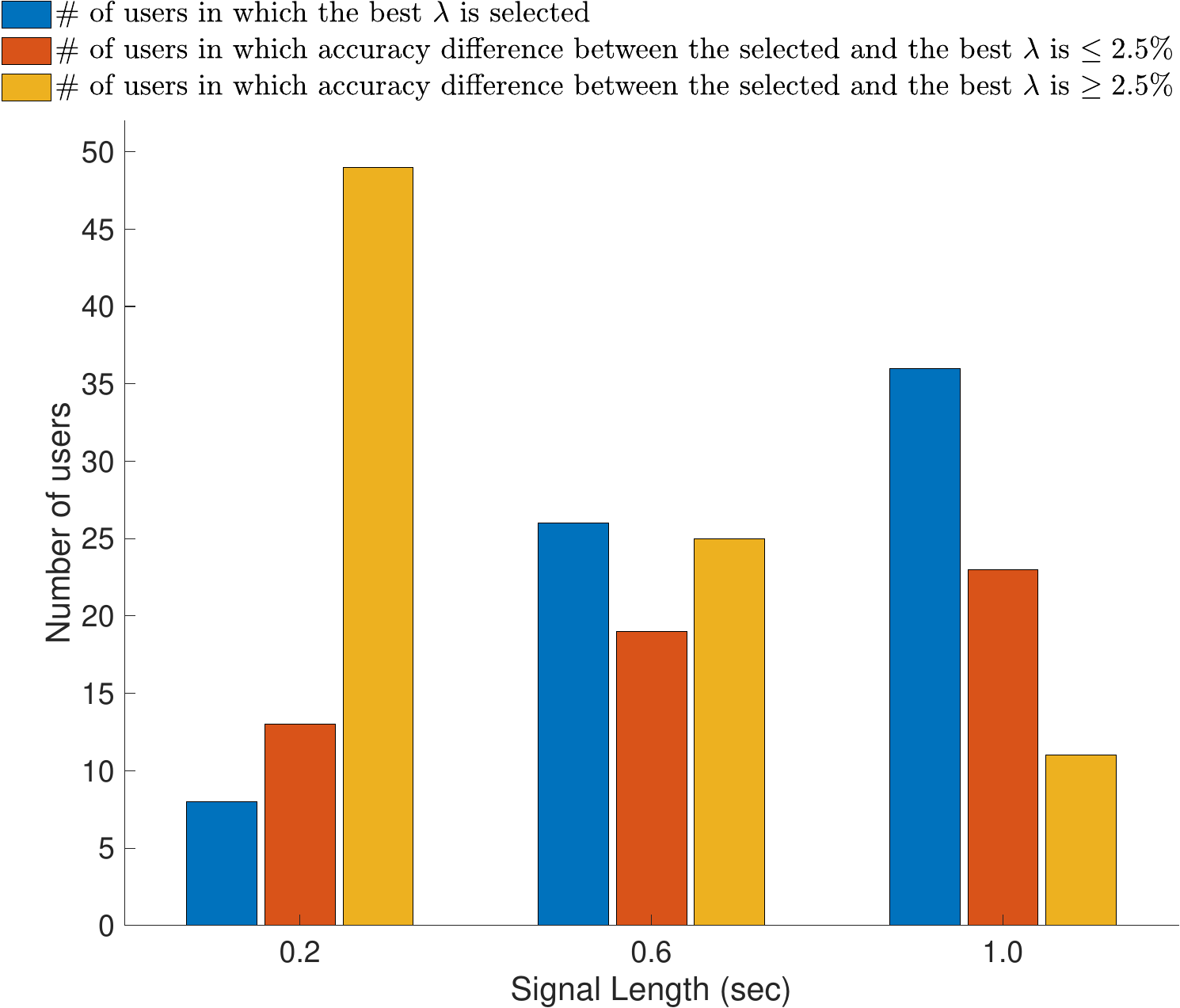}
        \caption{}
        \label{fig:best_lambda_beta}
    \end{subfigure}
    \caption{Our $\lambda$ selection is illustrated with the benchmark dataset \cite{benchmark_Dataset} (a) and the BETA dataset \cite{beta} (b) for three different signal lengths: $0.2$, $0.6$, and $1$ seconds. The graphs represent the distribution of users into three specific performance intervals of accuracy comparisons between the selected $\lambda$ and the best $\lambda$.}
\end{figure*}

\subsection{Datasets} \label{sec:datasets}
We use two large-scale datasets for performance evaluation: the benchmark \cite{benchmark_Dataset} and BETA \cite{beta}, which include SSVEP signals from $35$ and $70$ healthy participants, respectively, recorded with EEG from $64$ channels whilst performing the BCI spelling task. In the experimental setup, a visual representation of a $5 \times 8$ character matrix in the benchmark and a keyboard layout in BETA consisting of alphanumeric characters and symbols (e.g., blank space) was presented to the participants. In each dataset, there were a total of $40$ characters, each with a unique flickering frequency in the range from $8$ to $15.8$ Hz with a $0.2$ Hz increment, and at least a $0.5 \pi$ phase difference between adjacent characters. The participants were instructed to gaze toward a $0.5$-second visual cue (i.e., a red square) around the character to be spelled, followed by a flickering period and an offset (i.e., resting phase). Consequently, $6$-second SSVEP signals were recorded for each of the $40$ characters across $6$ blocks in the benchmark dataset, and $4$-second ($3$-second) SSVEP signals for the first $15$ participants (for the remaining $55$ participants) were recorded for each of the $40$ characters across $4$ blocks in BETA (including $0.5$ seconds of gaze shift and $0.5$ seconds for offset). Importantly, the SSVEP signals for the benchmark dataset were collected in a shielded lab environment, whereas those for BETA were collected outside the lab. Hence, BETA signals are of poorer quality, presenting a more challenging character recognition problem. More detailed information about the experimental setup and data collection can be found in \cite{benchmark_Dataset} for the benchmark dataset and in \cite{beta} for BETA.

\subsection{Results and Discussion}
The mean (across users) classification accuracy and  ITR with the standard errors are reported for the signal duration $T$ in the range $T \in \{0.2,0.4,\dots,1.0\}$. Also, $0.5$ seconds of gaze shift time that exists in both datasets is taken into account when calculating ITR. In all of the methods, a pre-determined set of $9$ channels (Pz, PO3, PO5, PO4, PO6, POz, O1, Oz, and O2) are used from the occipital and parietal regions. 

{The results presented in Figures \ref{fig:bench_res} and \ref{fig:beta_res} demonstrate that our method significantly outperforms competing approaches, achieving maximum ITRs of $201.15$ bits/min and $145.02$ bits/min on the benchmark and BETA datasets, respectively. These figures also present the pre-trained model performance to illustrate the improvement provided by our adaptation algorithm. The second-best performance is achieved by the OACCA method, which attains $183.92$ bits/min on the benchmark dataset and $141.80$ bits/min on BETA. Our method's superiority over other source-free domain adaptation methods becomes more pronounced with shorter signal durations (i.e., $0.2$, $0.4$, and $0.6$ seconds). For instance, at $T=0.4$ seconds, our method achieves $148.72$ bits/min (benchmark) and $100.81$ bits/min (BETA), while the best-performing competing source-free domain adaptation method reaches only $64.72$ bits/min (Adaptive-C3A on benchmark) and $39.50$ bits/min (OACCA on BETA). However, the domain generalization approach (ensemble of DNNs~\cite{our_ensemble}) shows better performance at $0.2$ seconds and comparable performance at $0.4$ seconds, particularly on the BETA dataset. This difference likely arises from the relatively poor signal quality under these conditions, which can negatively affect both pseudo-label quality and neighbor selection in our adaptation framework. These findings suggest that combining the ensemble of DNNs with our proposed adaptation algorithm may yield further improvements, highlighting a promising direction for future research. To our knowledge, these performance figures represent the best achieved results among user-independent methods (those requiring no calibration) in the literature (cf. Section \ref{sec:related_work}). Furthermore, our adaptation algorithm improves the maximum ITR of the pre-trained DNN by $62.81\%$ (from $123.54$ to $201.15$ bits/min) on the benchmark dataset and by $62.55\%$ (from $89.21$ to $145.02$ bits/min) on the BETA dataset. Appendix Figs.~\ref{fig:bench_box_plots} and \ref{fig:beta_box_plots} provide side-by-side box plots summarizing per-participant performance across signal lengths (medians and IQRs) for all methods, complementing the results reported here. Also note that the top-of-panel symbols denote the results of statistical analyses, specifically paired $t$-tests with Bonferroni thresholds, with further details provided in Sec.~\ref{sec:statistic_analysis}.}

\begin{table*}[t!]
\caption{The mean target identification accuracy (together with the standard errors) after adaptation, with fixed $\lambda$, across all $\lambda$ values, along with the best and selected $\lambda$ cases for signal lengths of 0.2, 0.6, and 1 seconds on the benchmark and BETA datasets. The last row presents the performance of the pre-trained model for comparison.}
\tiny
\centering
\begin{tabular}{@ {\extracolsep{4pt}}ccccccc}
\toprule   
{} & \multicolumn{6}{c}{Signal length (sec)}  \\
\cmidrule{2-7} 
{} & \multicolumn{3}{c}{Benchmark} &\multicolumn{3}{c}{BETA}  \\
\cmidrule{2-7} 

\centering 
 {$\lambda$} & 0.2 & 0.6 & 1.0 & 0.2 & 0.6 & 1.0 \\ 
 \midrule

{0.0 (Only $\mathcal{L}_{ll}$)} & 24.00 $\pm$ 2.63 &72.17 $\pm$ 4.52    &89.64 $\pm$ 3.20 &17.96 $\pm$ 1.38 & 54.03 $\pm$ 3.26 & 73.71 $\pm$ 2.70 \\
{0.2} & 23.17 $\pm$ 2.72 &70.26 $\pm$ 4.92    &91.10 $\pm$ 2.52 &18.91 $\pm$ 1.33 & 56.68 $\pm$ 3.12 & 75.20 $\pm$ 2.58 \\
{0.4} & 23.51 $\pm$ 2.72 &73.25 $\pm$ 4.44    &91.56 $\pm$ 2.25 &18.24 $\pm$ 1.43 & 56.62 $\pm$ 3.17 & 76.48 $\pm$ 2.36 \\
{0.6} & 22.85 $\pm$ 2.81 &73.20 $\pm$ 4.60    &91.93 $\pm$ 2.35 &18.17 $\pm$ 1.46 & 57.46 $\pm$ 3.24 & 76.45 $\pm$ 2.49 \\
{0.8} & 23.06 $\pm$ 2.85 &71.18 $\pm$ 5.09    &92.17 $\pm$ 2.23 &17.72 $\pm$ 1.35 & 57.54 $\pm$ 3.18 & 76.92 $\pm$ 2.36 \\
{1.0 (Only $\mathcal{L}_{sl}$)} & 22.38 $\pm$ 2.83 &69.58 $\pm$ 5.04    &90.85 $\pm$ 2.50 &18.18 $\pm$ 1.44 & 57.11 $\pm$ 3.23 & 76.48 $\pm$ 2.36 \\
\cdashline{1-7} 
{Best} & 26.44 $\pm$ 2.73 &76.48 $\pm$ 4.11    &93.02 $\pm$ 2.01 &22.06 $\pm$ 1.44 & 61.08 $\pm$ 2.99 & 78.54 $\pm$ 2.25 \\
{Selected} & 22.83 $\pm$ 2.95 &74.50 $\pm$ 4.57    &92.39 $\pm$ 2.07 &17.07 $\pm$ 1.43 & 58.57 $\pm$ 3.25 & 77.56 $\pm$ 2.36 \\
\midrule
{Pre-trained} & 21.20 $\pm$ 2.16 &51.19 $\pm$ 3.92    &70.61 $\pm$ 3.92 &20.21 $\pm$ 1.80 & 44.88 $\pm$ 3.52 & 56.58 $\pm$ 3.85 \\
\bottomrule
\end{tabular}
\label{table:diff_lambdas}
\end{table*}

Fig. \ref{fig:best_lambda_bench} and Fig. \ref{fig:best_lambda_beta} analyze our $\lambda$ selection. Recall that in our method, the pre-trained DNN is adapted to the target domain multiple times, each time with a different $\lambda$ chosen from $\Lambda=\{0, 0.2, 0.4, 0.6, 0.8, 1\}$. In hindsight, the best $\lambda$ (which is unknown in the process) is the one yielding the highest accuracy when tested with true labels at the end of the adaptation. In this analysis, we check the deviation between the accuracy performance of the best $\lambda$ and that of $\lambda$ selected according to the overall silhouette score  as explained in Section \ref{sec:methodology}. Three performance intervals of accuracy comparisons are considered: (i) no deviation (the best $\lambda$ is selected successfully), (ii) deviation within a margin of $2.5 \%$ (the best $\lambda$ is selected approximately), and (iii) out of margin deviation (the best $\lambda$ is missed). The reason for using $1/40 = 2.5\%$ as the margin threshold is that it is the chance level in the current classification problem. In the benchmark dataset ($35$ users in total), the best $\lambda$ is successfully selected for $10$, $19$ and $20$ users in the cases of $0.2$, $0.6$ and $1$ seconds of stimulation, respectively. In the BETA dataset ($70$ users in total), the best $\lambda$ is successfully selected for $8$, $26$ and $36$ users in the cases of $0.2$, $0.6$ and $1$ seconds of stimulation, respectively. Therefore, our $\lambda$ selection improves significantly with the increasing signal length $T$ (stimulation duration), because the signal-to-noise ratio (SNR) of the SSVEP harmonics increases with longer stimulation leading to more reliable silhouette clustering metric computations. The SNR level also explains having better $\lambda$ selection for the benchmark dataset compared to BETA as the BETA signals collected outside lab are of lower SNR and noisier.

\begin{figure*}[t!]
    \centering
    \begin{subfigure}[b]{0.45\textwidth}
        \centering
        \includegraphics[width=\textwidth]{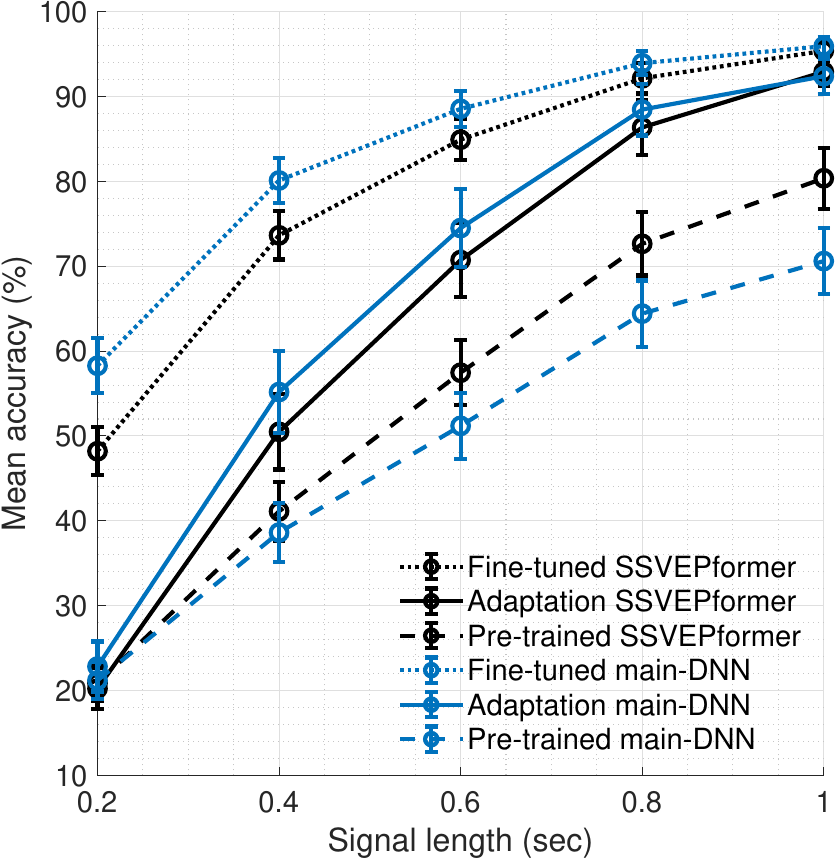}
        \caption{}
        \label{fig:ssvepformer_bench}
    \end{subfigure}
    \hspace{0.025\textwidth}
    \begin{subfigure}[b]{0.45\textwidth}
        \centering
        \includegraphics[width=\textwidth]{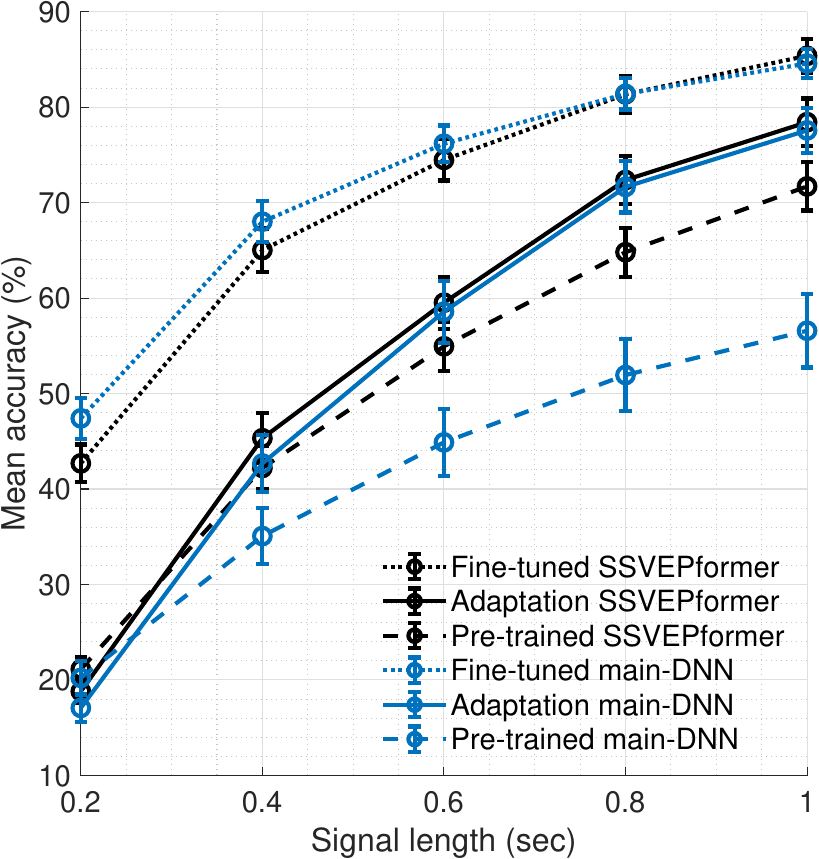}
        \caption{}
        \label{fig:ssvepformer_beta}
    \end{subfigure}
    \caption{The mean classification accuracy results are presented across all users in the datasets, together with the standard errors indicated by the bars. These results are for after fine-tuning with calibration data, adaptation using our approach, and the pre-trained versions, using two different DNN architectures. (a) The results for the benchmark dataset \cite{benchmark_Dataset}. (b) The results for the BETA dataset \cite{beta}.}
\end{figure*}

Table \ref{table:diff_lambdas} shows the performance results after adaptation using the main DNN \cite{ournetwork}, with fixed $\lambda$, across all $\lambda$ values, as well as the best and selected $\lambda$ cases. For comparison purposes, we also include the performance of the pre-trained model. When $\lambda=0$, the network is adapted using only the local-regularity term ($\mathcal{L}_{ll}$), while at $\lambda=1$, it is adapted using only the self-adaptation term ($\mathcal{L}_{sl}$). The superior performances after the adaptation compared to the pre-trained versions demonstrate the effectiveness of using both terms for adaptation, further highlighting the weighted combination as particularly more effective. Notably, among the fixed $\lambda$ values, the highest mean accuracy is never achieved by either $\lambda=0$ or $\lambda=1$. Additionally, the performance differences between all fixed $\lambda$ values and both the selected and best $\lambda$ cases indicate that the optimal $\lambda$ varies across different participants. The best $\lambda$ case establishes the practical maximum performance limit for our method. As shown in Fig. \ref{fig:best_lambda_bench} and Fig. \ref{fig:best_lambda_beta}, the results in this table further demonstrate that our $\lambda$ selection gets better as the signal length increases, with the performance gap between the best and selected cases decreasing.

Figures \ref{fig:ssvepformer_bench} and \ref{fig:ssvepformer_beta}, along with Table \ref{table:EAA}, present the results of our adaptation method using two other architectures: SSVEPformer \cite{ssvepformer} and the alternative DNN \cite{another_network}. These results further highlight the effectiveness of our adaptation approach and its generalizability across different DNN architectures. Fig. \ref{fig:ssvepformer_bench}, and Fig. \ref{fig:ssvepformer_beta}, present three sets of results, which are fine-tuning with calibration data, adaptation using our approach, and the pre-trained
versions, are presented for the main-DNN \cite{ournetwork} and the SSVEPformer \cite{ssvepformer} architecture. We obtain the fine-tuned performance results by fine-tuning the pre-trained models, treating some blocks of the user/test participant as calibration data, following a leave-one-block-out approach similar to \cite{ournetwork}. Specifically, we fine-tune the pre-trained architectures using 5 (or 3) blocks as calibration data and test on the remaining block, repeating this process 6 (or 4) times to evaluate each block in the benchmark (or BETA) dataset. These results compare the advantages and disadvantages of two state-of-the-art DNN architectures proposed for SSVEP-based BCIs. The main-DNN \cite{ournetwork} outperforms SSVEPformer \cite{ssvepformer} after fine-tuning with calibration data, whereas SSVEPformer performs better in the pre-trained versions. These results suggest that when sufficient data is available for adaptation (as in the benchmark dataset), the DNN architecture that performs better in its fine-tuned version also achieves superior performance after adaptation. However, when adaptation data is limited (as in the BETA dataset), using the DNN architecture that is better in its pre-trained version becomes more advantageous. Additionally, since the pre-trained SSVEPformer outperforms the pre-trained main-DNN on average, it can provide better initial pseudo-labels. To leverage the strengths of both architectures and assess the benefits of this performance difference in the pre-trained setting, we adapt the main-DNN at $T=0.8$ using the initial pseudo-labels generated by the SSVEPformer architecture. This approach increases the maximum ITR attained using the main-DNN on the benchmark dataset from $201.15$ bits/min to $202.54$ bits/min, and on the BETA dataset from $145.02$ bits/min to $149.31$ bits/min.

\begin{table}[t!]
\caption{The comparison of mean target identification accuracy (first rows) and mean ITR (second rows) for the alternative DNN before (global) and after adaptation.}
\centering
\begin{tabular}{@ {\extracolsep{4pt}}ccccc}
\toprule   
{} & \multicolumn{2}{c}{Benchmark} &\multicolumn{2}{c}{BETA}  \\
\cmidrule{2-3} 
\cmidrule{4-5} 
\centering 
 {Datasets} & Global & Adaptation 
 & Global & Adaptation \\ 
\cmidrule{1-3} 
\cmidrule{4-5} 
{} & 61.00  &76.18          &50.55 & 57.91 \\
{0.768 sec} & 116.15 &163.17    &87.35& 108.24 \\
\bottomrule
\end{tabular}
\label{table:EAA}
\end{table}

\begin{table}[t!]
\caption{ {Performance of the main DNN after adaptation when varying the number of target-domain blocks of unlabeled data used for adaptation at a signal length of $0.8$ seconds. For each dataset, the first row reports classification accuracy, and the second row reports ITR (bits/min). For the BETA dataset, results with 5 and 6 blocks are not applicable (N/A) since it contains only 4 blocks of data. }}
\centering
\begin{tabular}{@ {\extracolsep{4pt}}ccccccc}
\toprule   
\centering 
 {} & 1 block & 2 blocks  & 3 blocks & 4 blocks & 5 blocks & 6 blocks \\ 
 \midrule
{} & 81.06 & 80.18  &85.36 &86.36 & 87.04  & 88.46 \\
{Benchmark} & 177.01 & 173.92 &191.28 &194.71 & 197.46 & 201.15 \\
\midrule
{} & 62.31 & 64.23  &68.85 &71.64 &N/A & N/A \\
{BETA} & 118.35 & 123.79 &137.55 &145.02& N/A & N/A  \\

\bottomrule
\end{tabular}
\label{table:eventual}
\end{table}

\begin{table}[t!]
\caption{ {Ablation study on the parameters $\delta$ (threshold for determining neighborhood size in Equation \ref{eq:threshold_neighbour}) and $\beta$ (weight of the L2 regularization term in Equation \ref{eq:total_loss}). Default values are $\delta = 0.05$ and $\beta = 0.001$. When varying one parameter, all others are fixed at their default values.}}
\centering
\begin{tabular}{@{\extracolsep{4pt}}ccccccc}
\toprule   
\centering 
 {} & \multicolumn{3}{c}{$\delta$}  & \multicolumn{3}{c}{$\beta$}   \\ 
 \cmidrule{2-4}
 \cmidrule{5-7}
 {Datasets} & $0.01$  & $0.05$ & $0.1$ & $0.0005$  & $0.001$ & $0.005$ \\ 
 \cmidrule{1-4}
 \cmidrule{5-7}
{} & 87.48  &88.46 &88.21 & 88.37  & 88.46  & 87.33 \\
{Benchmark} & 198.56 &201.15 &200.07 & 200.88 & 201.15 & 198.25 \\
\cmidrule{1-4}
 \cmidrule{5-7}
{} & 71.64 & 71.64 &70.81 &71.04 & 71.64 &70.91\\
{BETA} & 145.29 &145.02 &142.80& 144.07 & 145.02 &143.43  \\
\bottomrule
\end{tabular}
\label{table:ablation}
\end{table}

In Table \ref{table:EAA}, the results are presented using the alternative DNN \cite{another_network} with a signal duration of $T=0.768$ seconds on both datasets. This alternative DNN originally is designed for the sample size of $256$, to align with its original design we want to keep the sample size to multiple of $64$ (its convolutional kernel sizes are $8$, $16$, $32$, and $64$). Also, we want to test on the signal duration as close as to $T=0.8$, where our main-DNN achieved the maximum ITR. Considering these factors, we selected a signal duration of $T=0.768$ seconds. To adopt this network to our approach, we have made some minor modifications, which are detailed in the Appendix. The results show that after adapting the alternative DNN with our algorithm its ITR performance is improved by $40.48\%$ on the benchmark and by $23.92\%$ on the BETA dataset. These improvements once again indicate the applicability of our algorithm in different settings. Note that the adaptation results for the SSVEPformer \cite{ssvepformer} and the alternative DNN architectures presented, in Fig. \ref{fig:ssvepformer_bench}, Fig. \ref{fig:ssvepformer_beta} and Table \ref{table:EAA}, are obtained with all hyperparameters kept the same as those used for our main-DNN in these experiments, so those results can be improved further by doing hyperparameter tuning. 

To assess the robustness of the proposed method under varying amounts of available target-domain data, we conduct experiments in which only a fraction of the unlabeled data from each user is used for adaptation rather than the entire set. The adapted DNN is subsequently evaluated on the full set of unlabeled target data. The signal duration is set to 0.8 seconds, as the maximum ITR is achieved at this length. Specifically, we vary the number of blocks used for adaptation in each dataset, with blocks for each participant selected at random. The results, presented in Table~\ref{table:eventual}, demonstrate that the proposed adaptation algorithm remains effective even when only a limited amount of the target data is available, with performance showing consistent improvement as the amount of adaptation data increases. {The only exception is the performance difference observed between one and two blocks of data on the benchmark dataset, which can be attributed to the random selection of blocks for adaptation.} The most notable gains occur when the number of adaptation blocks increases from two to three, suggesting that three blocks of data may be sufficient for effective adaptation. In particular, on the benchmark dataset, using three blocks for adaptation yields an ITR of $194.71$ bits/min (compared to the maximum of $201.15$ bits/min with six blocks), while on the BETA dataset, three blocks yield $137.55$ bits/min (compared to the maximum of $145.02$ bits/min with four blocks). For the one-block experiments, our $\lambda$ selection algorithm tends to favor networks adapted with higher $\lambda$ values (i.e., $0.8$ and $1.0$), as expected.

{Table \ref{table:ablation} presents the results ablation experiments. We test the sensitivity of our method to varying values of $\delta$ (the threshold for determining neighborhood size in Equation \ref{eq:threshold_neighbour}) and $\beta$ (the parameter controlling the weight of L2 regularization term in Equation \ref{eq:total_loss}). The defaults are $\delta=0.05$ and $\beta=0.001$. When we vary $\delta$ or $\beta$, others are kept constant at their defaults. The results suggest that our method shows limited sensitivity to the variation of these parameters.}

In addition to evaluating performance, we also assessed the computational efficiency of our adaptation method. On an Intel Core i7-10750 CPU, a single iteration takes approximately 30 seconds on average, and adaptation for one $\lambda$ takes around 10 iterations. {It is worth noting that the adaptation process can be accelerated through more efficient implementation strategies, such as vectorization and GPU-based execution. While we have not yet applied such optimizations, we recognize that they represent a promising direction to substantially reduce adaptation time.}

We also would like to point out a limitation of our method: it adapts the network in a batch setting, assuming that we have some accumulated unlabeled data from the new user. This assumption is reasonable for speller systems, where users are expected to engage with the system for extended periods. However, it may not apply to other SSVEP-based BCI systems with shorter usage times. If this assumption does not hold, the local-regularity term may have limited or no effect. Nonetheless, the network can still be adapted with limited data using the self-adaptation term alone. {Furthermore, our adaptation method currently operates with a fixed window length. Yet the optimal signal length may vary across users, and even across instances for the same user \cite{dynamic_window_reason}. For this reason, dynamic windowing strategies \cite{dynamic_window1,dynamic_window2,dynamic_window3} have been shown to improve performance under such variability. Incorporating dynamic windowing into our framework therefore represents a promising direction for future research.}

\subsection{Statistical Significance Analyses}\label{sec:statistic_analysis}
{For each $T\in \{0.2, 0.4, 0.6, 0.8, 1\}$, we conduct $4$ paired $t$-tests, pairing our proposed adaptation method with the compared methods in Fig. \ref{fig:bench_res} and Fig. \ref{fig:beta_res}, as similarly done in \cite{ournetwork, our_ensemble}. Unadjusted p-values are reported, and we call the observed difference as ``statistically significant'' (*) if the p-value is less than $\frac{0.05}{4}$ and ``statistically highly significant'' (**) if the p-value is less than $\frac{0.05}{4\times 5}$. For the statistically significant case, single Bonferroni correction is applied by dividing $0.05$ by $4$, since for each $T$ there are $4$ comparisons. And for the statistically highly significant case, we apply double Bonferroni correction by $20$, since across all methods and $T$ choices there are $4\times5=20$ comparisons. Alongside $p$, we report Cohen’s $d_z$ for paired comparisons with 95\% confidence intervals, using a robust estimator with 1000 bootstrap resamples \cite{robust_cohen}. The marker $^{\downarrow}$ denotes effects favoring the comparator.

\textbf{In the case of the benchmark dataset:} In terms of the accuracy (Fig. \ref{fig:bench_res}), the least significant difference between our method (Algorithm \ref{alg:adaptation}) and the compared methods is observed with (i) ensemble of DNNs ($p=0.966$; $d_z=-0.035$ [$-0.192$, $0.11 $]) for $T=0.2$, (ii) ensemble of DNNs (**$p=4.83 \times 10^{-4}$; $d_z=0.396$ [$0.142$, $0.776 $]) for $T=0.4$, (iii) OACCA (*$p=0.94 \times 10^{-2}$; $d_z=0.42$ [$0.134$, $0.717 $]) for $T=0.6$, (iv) OACCA (*$p=0.0104$; $d_z=0.406$ [$0.14$, $0.784 $]) for $T=0.8$, (v) OACCA ($p=0.0702$; $d_z=0.182$ [$-0.062$, $0.404 $]) for $T=1.0$. At $T=1.0$ and $T=0.2$, no significant differences are observed compared to OACCA and the ensemble of DNNs, respectively, but highly significant differences (**) exist with all other methods. In terms of the ITR (Fig. \ref{fig:bench_res}), the least significant difference between our method (Algorithm \ref{alg:adaptation}) and the compared methods is observed with (i) ensemble of DNNs ($p=0.263$; $d_z=0.026$ [$-0.15$, $0.17 $]) for $T=0.2$, (ii) ensemble of DNNs (**$p=1.8 \times 10^{-5}$; $d_z=0.463$ [$0.244$, $0.827 $]) for $T=0.4$, (iii) OACCA (*$p=0.32 \times 10^{-2}$; $d_z=0.407$ [$0.137$, $0.692 $]) for $T=0.6$, (iv) OACCA (*$p=0.44 \times 10^{-2}$; $d_z=0.38$ [$0.141$, $0.697 $]) for $T=0.8$, and (v) OACCA ($p=0.092$; $d_z=0.154$ [$-0.068$, $0.369 $]) for $T=1$. For $T=1.0$, the difference with OACCA is not significant; but highly significant (**) with all the others.

\textbf{In the case of the BETA dataset:} In terms of the accuracy (Fig.~\ref{fig:beta_res}), the least significant difference between our method (Algorithm~\ref{alg:adaptation}) and the compared methods is observed with (i) ensemble of DNNs ($^{\downarrow}$**$p=2.82\times10^{-8}$; $d_z=-0.469$ [$-0.685$, $-0.312$]) for $T=0.2$, (ii) ensemble of DNNs ($p=0.28$; $d_z=0.052$ [$-0.061$, $0.170$]) for $T=0.4$, (iii) OACCA (*$p=5.1\times10^{-3}$; $d_z=0.211$ [$0.038$, $0.372$]) for $T=0.6$, (iv) OACCA ($p=0.021$; $d_z=0.098$ [$-0.027$, $0.202$]) for $T=0.8$, and (v) OACCA ($p=0.96$; $d_z=-0.041$ [$-0.210$, $0.085$]) for $T=1.0$. At $T=0.8$ and $T=1.0$, the differences with OACCA are not significant, but highly significant (**) with all other methods. At $T=0.2$, the effect favors the ensemble of DNNs ($d_z<0$), while at $T=0.4$ the difference is not significant; in both cases, however, our method remains highly significant (**) against all other comparators. In terms of the ITR (Fig. \ref{fig:beta_res}), the least significant difference between our method (Algorithm \ref{alg:adaptation}) and the compared methods is observed with (i) ensemble of DNNs ($^{\downarrow}$**$p=4.53\times10^{-6}$; $d_z=-0.446$ [$-0.676$, $-0.26$]) for $T=0.2$, (ii) ensemble of DNNs (*$p=0.101$; $d_z=0.103$ [$0.001$, $0.245$]) for $T=0.4$, (iii) OACCA (*$p=0.28 \times 10^{-2}$; $d_z=0.214$ [$0.006$, $0.382$]) for $T=0.6$, (iv) OACCA ($p=0.025$; $d_z=0.09$ [$-0.024$, $0.191$]) for $T=0.8$, and (v) OACCA ($p=0.76$; $d_z=-0.043$ [$-0.2$, $0.069$]) for $T=1$. For $T=0.8$ and $T=1.0$, the differences with OACCA are not significant, but highly significant (**) with all the others. At $T=0.2$, the effect favors the ensemble of DNNs ($d_z<0$), but the differences are highly significant (**) with all the other methods.}

\section{Conclusion} \label{sec:CO}
In this study, we proposed a source free domain adaptation method for SSVEP-based BCIs. Our method adapts a DNN architecture (pre-trained with labeled data from source domains) to a new user (target domain) using only unlabeled target data by minimizing our proposed novel loss function. There are two main terms in the proposed loss function: self-adaptation and local-regularity. The self-adaptation term utilizes the pseudo-label strategy. The local-regularity term takes advantage of the data structure and forces the DNN to give similar labels to neighboured instances. We experimented with commonly used and publicly available two large scale datasets, which are the benchmark and BETA datasets, using three different DNN architectures. Our adaptation method consistently improves the performance of the pre-trained DNN across different architectures and outperforms all the compared alternatives in terms of accuracy and information transfer rate, achieving to the best of our knowledge the best performance within its own category of calibration-free methods. In our design, we specifically prioritized the user comfort by not requiring any calibration or preparation process and allowed an immediate system start. Hence, our method can be directly used in plug-and-play manner. In this sense, we believe that the presented work strongly promotes the spread of SSVEP-based BCI systems in daily life.

\section{Acknowledgements}
This work was supported in part by The Scientific and Technological Research Council of Turkiye (TUBITAK) under Contract 121E452. We thank P. Ciftcioglu and G. Coskun for their contributions in our preliminary experiments.

\newpage
\appendix
\renewcommand{\thetable}{A\arabic{table}}
\setcounter{table}{0}

\renewcommand{\thefigure}{A\arabic{figure}}
\setcounter{figure}{0}

\section*{Appendix}
\begin{figure}[t!]
    \centering
    \begin{subfigure}[b]{0.99\textwidth}
        \centering
        \includegraphics[width=\textwidth]{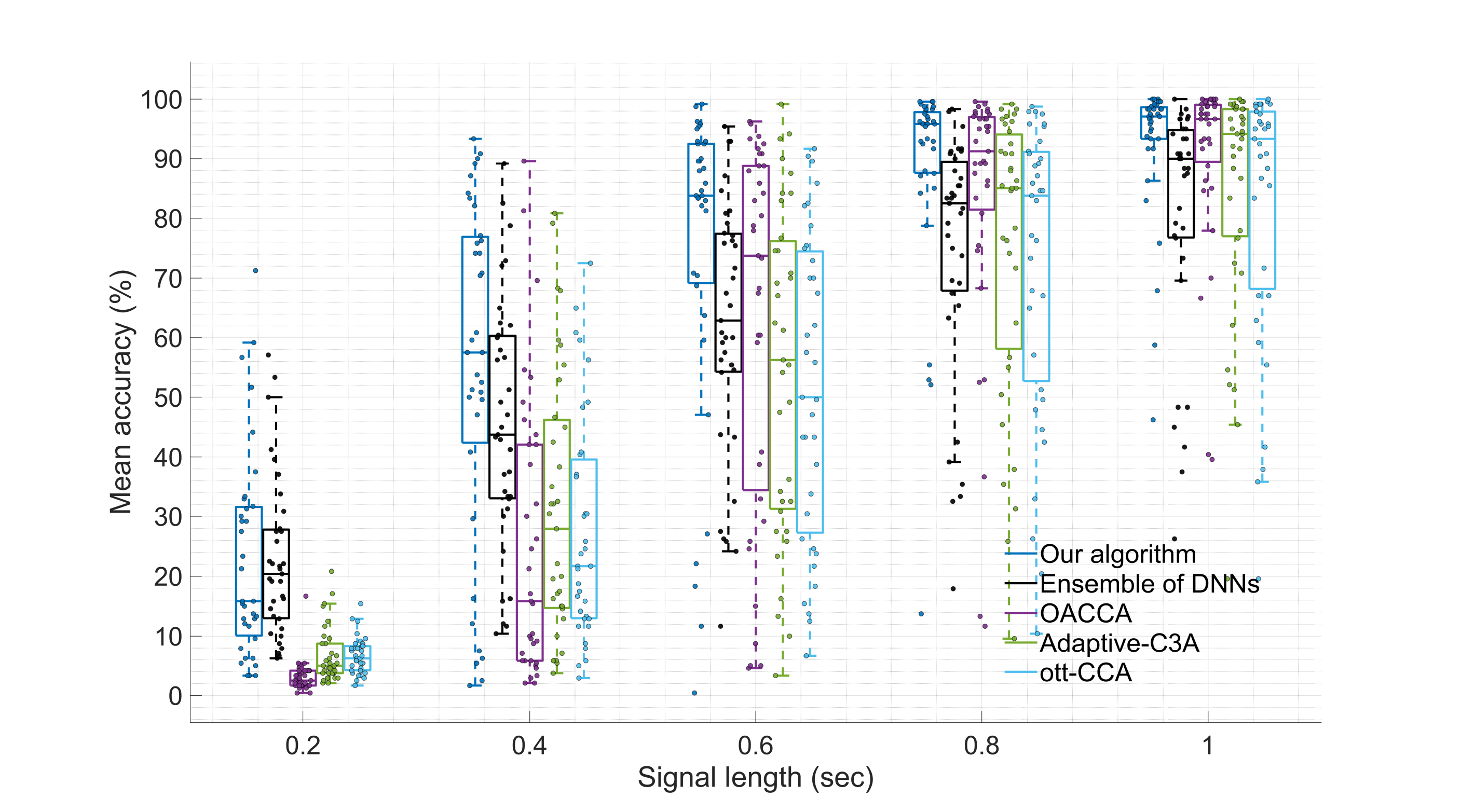}
        \caption{}
        \label{fig:bench_box_acc}
    \end{subfigure}
    \hspace{0.05\textwidth}
    \begin{subfigure}[b]{0.99\textwidth}
        \centering
        \includegraphics[width=\textwidth]{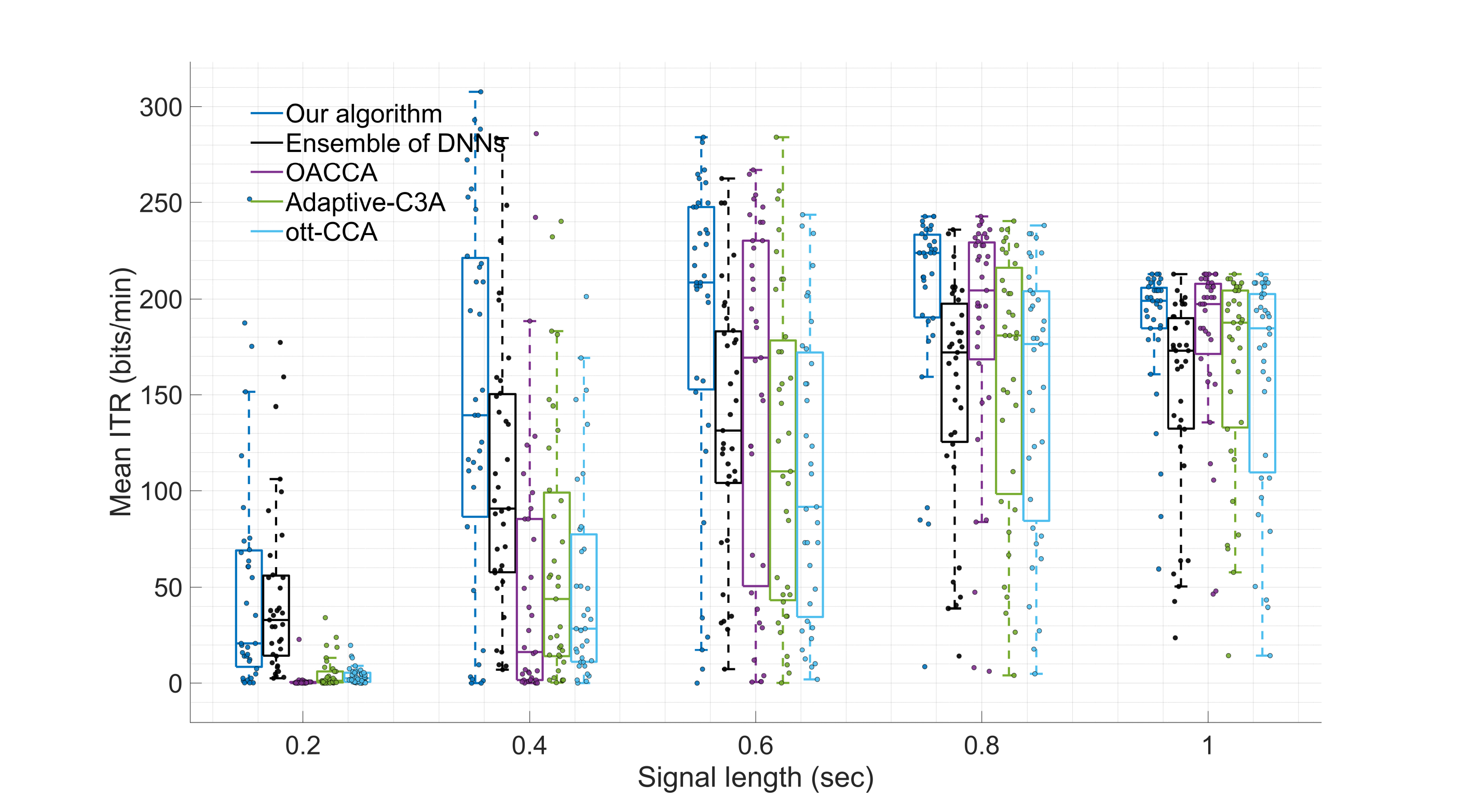}
        \caption{}
        \label{fig:bench_box_itr}
    \end{subfigure}
    \caption{For each signal length $T\in\{0.2,0.4,0.6,0.8,1.0\}\,$, side-by-side boxplots show per-participant performance for each method. Boxes depict the interquartile range (IQR; 25th–75th percentiles) with the median line; whiskers extend to $1.5\times$ IQR; jittered points indicate individual participants. (a) Mean accuracy. (b) Mean information transfer rate (ITR). Results are shown for the benchmark dataset.}
    \label{fig:bench_box_plots}
\end{figure}

\begin{figure}[t!]
    \centering
    \begin{subfigure}[b]{0.99\textwidth}
        \centering
        \includegraphics[width=\textwidth]{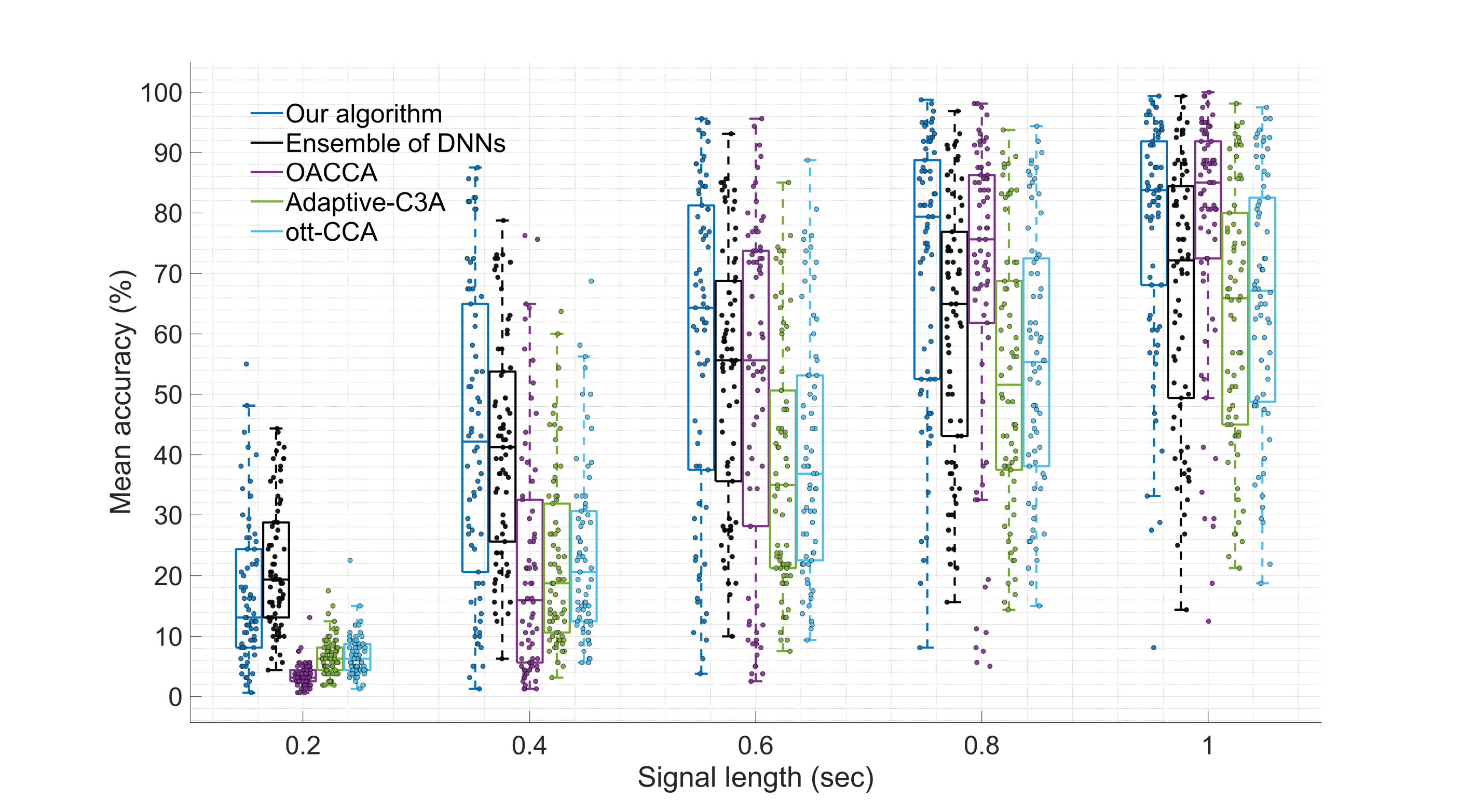}
        \caption{}
        \label{fig:beta_box_acc}
    \end{subfigure}
    \hspace{0.05\textwidth}
    \begin{subfigure}[b]{0.99\textwidth}
        \centering
        \includegraphics[width=\textwidth]{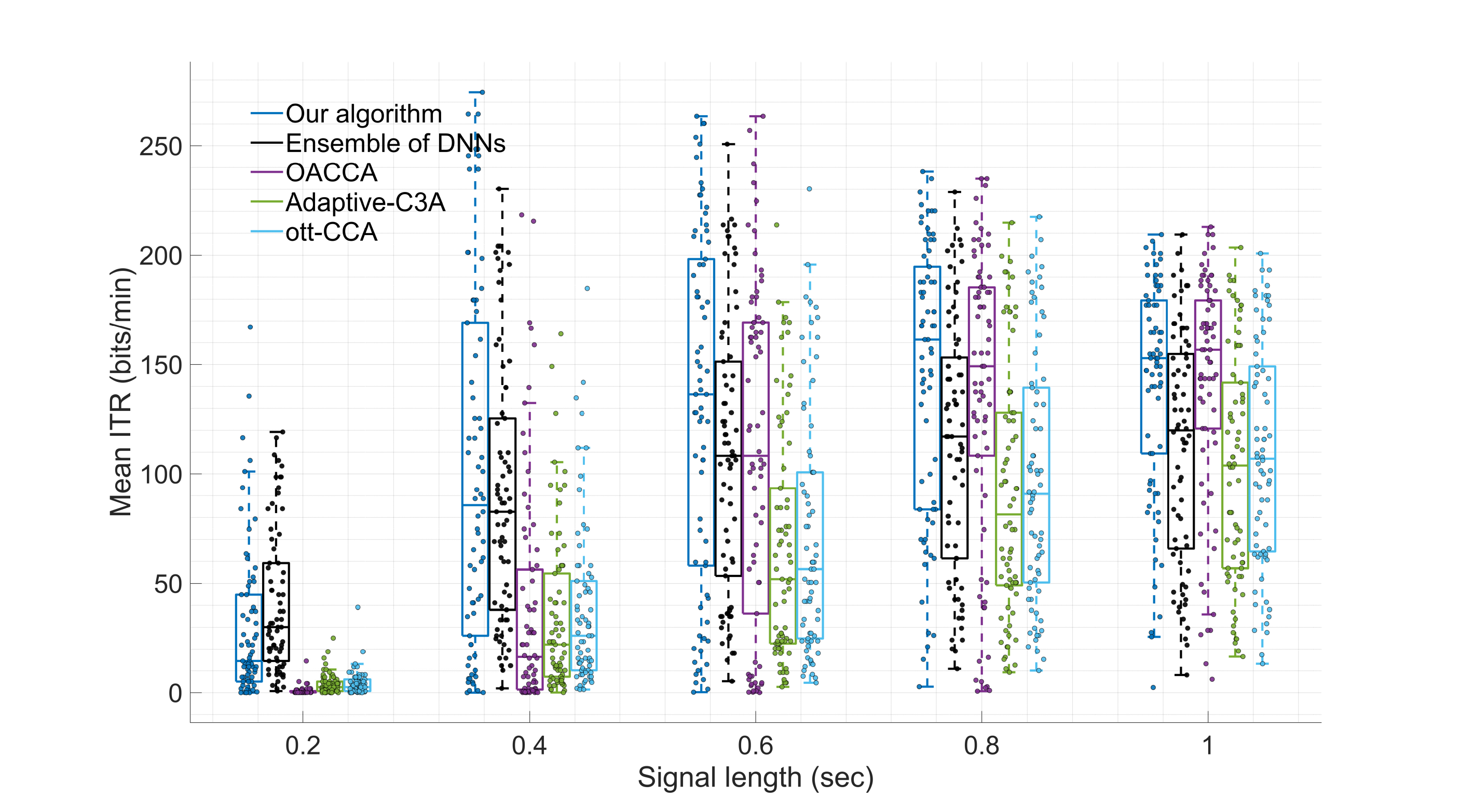}
        \caption{}
        \label{fig:beta_box_itr}
    \end{subfigure}
    \caption{For each signal length $T\in\{0.2,0.4,0.6,0.8,1.0\}\,$, side-by-side boxplots show per-participant performance for each method. Boxes depict the interquartile range (IQR; 25th–75th percentiles) with the median line; whiskers extend to $1.5\times$ IQR; jittered points indicate individual participants. (a) Mean accuracy. (b) Mean information transfer rate (ITR). Results are shown for the BETA dataset.}
    \label{fig:beta_box_plots}
\end{figure}

In this section, we explain the implementation details of the main DNN architecture \cite{ournetwork}, {the SSVEPformer architecture \cite{ssvepformer}, } the alternative DNN architecture \cite{another_network}, the neighbor selection as well as confidence of the instances. Also, we provide outline of our adaptation algorithm in Algorithm \ref{alg:adaptation}. {Furthermore, Figs.~\ref{fig:bench_box_plots} and \ref{fig:beta_box_plots} present side-by-side box plots that summarize per-participant performance across signal lengths, reporting medians and interquartile ranges (IQRs) for all methods.} Finally, we present how the main DNN \cite{ournetwork} performed after adaptation using different fixed values of $\lambda$, including both the best and selected $\lambda$ cases. Our code is available at 
\url{https://github.com/osmanberke/SFDA-SSVEP-BCI}

\subsection{The DNN Architecture}
The main DNN \cite{ournetwork} consists of four convolutional and one fully connected layers. The first convolutional layer combines the sub-bands, whereas the second combines the channels. The third and fourth convolutional layers extract the features by processing the EEG signals in time and the last fully connected layer obtains the spelled character prediction. During training, dropout \cite{dropout_paper} is applied after the second, third, and fourth layers with probabilities of 0.1, 0.1, and 0.95, respectively. The convolutional layers have fixed number of parameters for all signal lengths, while the number of parameters in the final fully connected layer scales linearly with signal length. We use the suggested setting in the original paper \cite{ournetwork}, namely we use 3 sub-bands and 120 many channel combinations. Also note that, in the experiments, we have 9 channels and 40 classes. With this setting, and using the formulation given \cite{ournetwork}, the total number of weights can be found as: $3$ (first layer)+ $9\times120$ (second layer) + $2\times120\times120$ (third layer) + $10\times120\times120$ (fourth layer) + $40\times120\frac{N_T}{2}$. For $T=0.8$ (resulting in $N_T = 200$), the total number of weights is $653,883$. Including the bias terms, the total parameter count is $654,284$.

This DNN architecture achieves the best reported accuracy and ITR results on the benchmark \cite{benchmark_Dataset} and BETA \cite{beta} datasets to date, when it is trained in a supervised setting with the calibration data. We refer the interested reader to \cite{ournetwork} for the full description.

\subsection{SSVEPformer Architecture}
The SSVEPformer architecture adapts the transformer model \cite{transformer_attention} by replacing its traditional attention mechanism with convolutional operations. The model's input is derived from the complex spectrum of EEG signals, obtained by concatenating the real and imaginary components of the channel-wise Fast Fourier Transform (FFT). The frequency resolution of FFT is kept at 0.2 Hz. To achieve this resolution, 5 seconds of signal length is required, so zero-padding is used when necessary. Instead of the full spectrum, only frequencies between 8 Hz and 64 Hz are used. This model consists of three core blocks: the channel combination, the encoder, and the multi-layer perceptron (MLP) head. The channel combination block combines the channels and applies layer normalization \cite{ba2016layernormalization} followed by non-linear activation. The encoder block consists of two sub-encoders for feature extraction. Each sub-encoder utilizes convolutional and fully connected layers while incorporating residual connections, layer normalization, and non-linear activations. Finally, the MLP head, consisting of two fully connected layers, performs the final character prediction. The dropout \cite{dropout_paper} layer, with a probability of 0.5, is applied before each fully connected layer in the MLP head block, as well as at the end of each sub-encoder and the channel combination block. In this architecture, the number of parameters are the same for all signal lengths up to 5 seconds, due to the zero-padding applied to maintain an FFT resolution of 0.2 Hz. The total number of parameters is 3,086,024. We choose not to provide the parameter count per layer due to the high number of layers in the model. This architecture does not incorporate sub-bands; therefore, we use a single band-pass filter for preprocessing when employing this architecture. Notably, SSVEPformer has an extended variant, FB-SSVEPformer \cite{ssvepformer}, which supports sub-band processing at the cost of increasing the number of parameters proportional to the number of sub-bands used. Due to computational constraints, we choose to use SSVEPformer rather than FB-SSVEPformer.

The SSVEPformer architecture has demonstrated good performance in domain generalization scenario, achieving high accuracy and ITR. For a comprehensive description of the architecture, readers are directed to the original paper \cite{ssvepformer}.

\begin{table}[t!]
\caption{{Glossary of parameters and their default values.}}
\centering
\begin{tabularx}{\textwidth}{L{3cm} X}
\toprule
Parameters & Definitions \\
\midrule
$\lambda$ & Controls the relative contributions of the loss terms $\mathcal{L}_{sl}$ and $\mathcal{L}_{ll}$. Its value is adaptively selected for each new user (Equation \ref{eq:total_loss}). \\
$\beta$ & Controls the strength of the L2 regularization term (Equation \ref{eq:total_loss}). Default: $\beta=0.001$. \\
$\Lambda$ & Set of candidate values for $\lambda$: $\Lambda=\{0,\,0.2,\,\dots,\,1\}$. \\
$k_i$ & Number of neighbors assigned to instance $i$. Determined dynamically per instance (Equation \ref{eq:threshold_neighbour}). \\
$\delta$ & Threshold used to determine the number of neighbors (Equation \ref{eq:threshold_neighbour}). Default: $\delta=0.05$. \\
$B$ & Termination criterion: if the clustering score fails to improve over $B$ consecutive iterations, adaptation is stopped. Default: $B=3$. \\
$J$ & Number of epochs performed in each iteration. Default: $J=50$. \\
\bottomrule
\end{tabularx}
\label{table:parameters}
\end{table}

\begin{algorithm}[t!]
   \centering
   \caption{Source-Free Adaptation of DNN $f_{\matr w}$}
   \label{alg:adaptation}
\begin{algorithmic}[1]
   \STATE \textbf{Input:} Unlabeled data instances $\{(\matr{x}_i)\}_{i=1}^{N}$ from the user (target domain) and pre-trained model's weights ${\matr w^{(0)}}$.
   \STATE \textbf{Parameters:} Learning rate $\alpha=0.0001$, L2 regularization weight $\beta=0.001$, maximum number of trials $B=3$ and epochs $J=50$, and the set of candidate $\lambda \in \Lambda=\{0,0.2,\cdots,1\}$.
   \STATE Let the previous iteration $t-1$ be denoted by $t'=t-1$.
   \FOR{$\lambda \in \Delta$}
   \STATE {Calculate the initial clustering score   $m^{(0)}(\lambda)$}.
   \STATE {Set iteration $t=1$ and trial $b=0$.}
   \REPEAT   
   \FOR{$j=0$ \textbf{to} $J-1$}
   \STATE Apply gradient descent to update: \\ ${\matr w^{(t' + \frac{j+1}{J})}} = {\matr w^{(t' + \frac{j}{J})}} - \alpha \times   \nabla_{\matr w^{(t' + \frac{j}{J})}}\mathcal{L}_{{total}}^{(t' + \frac{j}{J})}(\lambda)$.
   \ENDFOR
   \STATE Calculate the clustering score  $m^{(t)}(\lambda)$.
   \IF{$m^{(t)}(\lambda) > m^{(t')}(\lambda)$}
        \STATE Calculate the new predictions $\hat{y}_i^{(t)}$ for all $i$.
        \STATE $t\leftarrow t+1$.
        \STATE Set trial $b=0$.
   \ELSE 
        \STATE $b\leftarrow b+1$
   \ENDIF
   \UNTIL{$b=B$ (terminal).}
   \ENDFOR
   \STATE $\lambda_{\text{max}} = \argmax_{\lambda \in \Delta} m^{(t)}(\lambda) $  
   \STATE \textbf{Return:} $\matr{w}^{(t)}({\lambda_{\text{max}}})$
\end{algorithmic}
\end{algorithm}

\subsection{The alternative DNN Architecture}
The alternative DNN, proposed in \cite{another_network}, originally consists of four convolutional blocks and one fully connected layer. Each block contains standard convolutional and batch normalization layers followed by an activation function. The first two blocks perform convolutions across both channels and time, while the last two blocks perform convolutions only across time. The convolutional kernel sizes of blocks in time are $8$, $16$, $32$ and $64$, respectively. Each block pads the input signal to keep the same output dimension as the input dimension. However, this padding operation prevents learning the channel combination necessary for the neighbor selection detailed in the following section. Therefore, the main modification we make to this DNN architecture is that we remove the padding operation at the first block and convolve the signal only across the channels. Additionally, we reduce the number of convolutional layers in each block by half to expedite training. Lastly, since the original version of this DNN does not incorporate sub-bands, we utilize only one band-pass filter for preprocessing. {After the modifications, this architecture contains a total of $853,256$ parameters at $T=0.768$.} We refer the interested reader to original paper of this DNN architecture \cite{another_network} for the full description.

\subsection{Neighbor Selection}
As the EEG instances $\matr{x}_i$'s are multidimensional, the correlation coefficient between two given instances cannot be calculated regularly by $\rho(\matr{a},\matr{b})=\frac{\matr{a}\matr{b}'}{||\matr{a}||||\matr{b}||}$ (assuming that the signals are zero mean row vectors, and $\matr{b}'$ is the transpose of $\matr{b}$). We calculate this correlation coefficient after reducing the dimension by combining the channels of the instances with the channel combination weight $\matr{w}_c^{(*)}(\lambda) \in \mathbb{R}^{C \times 1}$ as follows:  

\begin{equation}
    \rho(\matr{w}_c^{(*)}{'}(\lambda)\matr{x}_i,\matr{w}_c^{(*)}{'}(\lambda)\matr{x}_j), \label{eq:real-corr}
\end{equation}
where $\matr{w}_c^{(*)}{'}(\lambda)$ is the transpose of $\matr{w}_c^{(*)}(\lambda)$ and $\lambda$ is the weighting of the two terms in the final loss. Hence, all the equations in main text containing the correlation coefficient are dependent on the channel combination weights $\matr{w}_c^{(*)}(\lambda)$, including the distance term $d(\matr{x}_j,\matr{x}_i)$ in \eqref{eq:silhouette}. Note also that, with a slight abuse of notation, the correlation coefficient notation $\rho(\matr{x}_i,\matr{x}_j)$ in main text in fact refers to the computation after combining the channels as in \eqref{eq:real-corr}. Consequently, the silhouette score of the instance $m_i(\lambda)$ and the overall silhouette score $m(\lambda)$ become dependent of $\matr{w}_c^{(*)}(\lambda)$, which we denote in this section by $m_i(\lambda,\matr{w}_c^{(*)}(\lambda))$, $m(\lambda,\matr{w}_c^{(*)}(\lambda))$, respectively, but do not show it in the main text for simplicity in exposition.

On the other hand, the channel combination $\matr{w}_c^{(*)}(\lambda)$ is selected to maximize the overall silhouette score for the given predictions. The reason behind this selection is that the channel combination that maximizes the metric performance (i.e. silhouette score) is intuitively expected to describe the user's spatial characteristics well. However, since the direct maximization of this score is intractable, we follow a different strategy, and select the channel combination $\matr{w}_c^{(*)}(\lambda)$ that maximizes the overall silhouette score $m(\lambda,\matr{w}_c^{(*)}(\lambda))$ among the finite set of channel combination weights from the channel combination layer (e.g. second layer in \cite{ournetwork}) of the DNN as

\begin{equation*}
        {\matr{w}_c^{(*)}}(\lambda) = \argmax_{\matr{w}_c(\lambda) \in \{\matr{w}_c(\lambda)\}_j} 
    \frac{1}{N}\sum_{i=1}^{N} m_i(\lambda,\matr{w}_c(\lambda)).
     \label{eq:best_channel_comb}
\end{equation*}
Here, we stress that the variables actually change across (and so are further dependent on) iterations in our Algorithm \ref{alg:adaptation}. This, to be more precise, requires a superscript $^{(t)}$ in notation but we dropped it for simplicity here and in main text as well.

When the main DNN architecture \cite{ournetwork} is used, the weights in the channel combination layer are learned/tuned for the input whose sub-bands of harmonics (generated in the preprocessing step) are combined in the preceding DNN layer. Therefore, to effectively use the selected channel combination weights and to take advantage of the preprocessing step in the neighbors and silhouette score calculation, we first preprocess the original EEG input to generate the sub-bands of harmonics. Then, we combine the sub-bands with the weights from the sub-band combination layer (i.e, first layer in \cite{ournetwork}), and then produce $\matr{x}_i$ that is used in the calculation of the silhouette score as well as in the determination of the instances' neighbors. Note also that the network weights are updated over iterations, so the channel and sub-band combination weights are also updated, which affect the calculation of the neighbors of instances. Hence, at each iteration $t$, the neighbors of instances must be recalculated: firstly, the channel combination is selected to maximize the silhouette score $m^{(t)}(\lambda,\matr{w}_c^{(t*)}(\lambda))$ for the given predictions, then, the neighbors of instances are determined with the selected channel combination.

\subsection{Instance Confidence}
In the loss term, we use the pseudo-labels of the instances having the positive silhouette score as explained in main text. However, with this strategy, there is a probability that either the self-adaptation term or the local-regularity term cease to be applicable for an instance $\matr{x}_i$ at any iteration $t$. If the instance itself has the negative silhouette score (i.e., $m_i^{(t)}(\lambda,\matr{w}_c^{(*)}) < 0$  ), the self-adaptation term becomes inapplicable; and if all the determined instance's neighbors have the negative silhouette score (i.e., $m_j^{(t)}(\lambda,\matr{w}_c^{(*)}) < 0$, $\forall j \in I_i^{(t)}$), the local-regularity term becomes inapplicable. Therefore, to enable the instances to contribute equally to the loss term; if one of the terms becomes inapplicable for the instance $x_i$, we update the $\lambda$ term for that instance as follows:

\begin{equation*}
\begin{split}
    \mathcal{L}^{(t)}_{{total}} &= \frac{-1}{N}\sum_{i=1}^{N}\lambda_i^{(t)}\log(\matr{s}_{i,\hat{y}_i^{(t)}}^{(t)})\\ &+\frac{-1}{N}\sum_{i=1}^{N}[\frac{1-\lambda_i^{(t)}}{k_i}\sum_{i=1}^{k_i} \log(\matr{s}_{i,\hat{y}_{I_i(j)}^{(t)}}^{(t)})],
    \label{eq:positive_loss}
\end{split}
\end{equation*}
where $\lambda_i^{(t)}$ is the instance specific $\lambda$ term at iteration $t$ and it equals to the global $\lambda$, if the both loss terms (self-adaptation and local-regularity) are applicable; and it equals to $0$ or $1$, if one of the term is inapplicable:

\begin{equation*}
  \lambda_i^{(t)} =
    \begin{cases}
      1 & \text{if $m_j^{(t)}(\lambda,\matr{w}_c^{(*)}) < 0$, $\forall j \in I_i^{(t)}$}\\
      0 & \text{if $m_i^{(t)}(\lambda,\matr{w}_c^{(*)}) < 0$}\\
      \lambda & \text{otherwise}
    \end{cases}.      
\end{equation*}

Even both terms may become inapplicable for some instances, then we do not use those instances in the loss.

\subsection{Initial Predictions}
The performance of the initial DNN model, which is pre-trained with data from source domains, may not be satisfactory for some target domains, especially when a majority of source domains' data statistics are much different from those of the target domain. In such cases, the completely training-free algorithms, such as FBCCA, might give better (more accurate) initial predictions. For this reason, we get initial predictions from either the pre-trained initial model or the FBCCA method. We choose the one having an initial higher silhouette score.

\section*{References}

\bibliographystyle{ieeetr}
\bibliography{references}

\end{document}